%% file: main.tex
\documentclass[10pt,twocolumn,letterpaper]{article}

\usepackage{style}
\usepackage{times}
\usepackage{epsfig}
\usepackage{graphicx}
\usepackage{amsmath}
\usepackage{amssymb}
\usepackage{placeins}
\usepackage{subcaption}
\usepackage{tabularx}
\usepackage{multirow}
\usepackage{hhline}
\usepackage{arydshln}
\usepackage{graphbox}
\usepackage[export]{adjustbox}
\usepackage{enumitem}

\newcommand{\fig}[1]{Figure~\ref{fig:#1}}

\newcommand{\tab}[1]{Table~\ref{tab:#1}}

\newcommand{\eq}[1]{(\ref{eq:#1})}

\newcommand{\e}{e}

\newcolumntype{Y}{>{\centering\arraybackslash}X}

\usepackage[pagebackref=true,breaklinks=true,letterpaper=true,colorlinks,bookmarks=false]{hyperref}
\usepackage[nameinlink,capitalise]{cleveref}

\stylefinalcopy

\begin{document}

\title{Few-Shot Adversarial Learning of Realistic Neural Talking Head Models\vspace{-0.28cm}}

\author{
Egor Zakharov$^{1,2}$ \quad Aliaksandra Shysheya$^{1,2}$ \quad Egor Burkov$^{1,2}$ \quad Victor Lempitsky$^{1,2}$
\vspace{1.5mm}\\
$^1$Samsung AI Center, Moscow \quad $^2$Skolkovo Institute of Science and Technology
}

\input{figures/teaser.tex}

\begin{abstract}
Several recent works have shown how highly realistic human head images can be obtained by training convolutional neural networks to generate them. In order to create a personalized talking head model, these works require training on a large dataset of images of a single person. However, in many practical scenarios, such personalized talking head models need to be learned from a few image views of a person, potentially even a single image. Here, we present a system with such few-shot capability. It performs lengthy meta-learning on a large dataset of videos, and after that is able to frame few- and one-shot learning of neural talking head models of previously unseen people as adversarial training problems with high capacity generators and discriminators. Crucially, the system is able to initialize the parameters of both the generator and the discriminator in a person-specific way, so that training can be based on just a few images and done quickly, despite the need to tune tens of millions of parameters. We show that such an approach is able to learn highly realistic and personalized talking head models of new people and even portrait paintings.
\end{abstract}

\input{intro}
\input{related}
\input{method}
\input{experiments}
\input{conclusion}

{\small
\bibliographystyle{ieee}
\bibliography{refs}
}

\appendix
\input{suppmat}

\end{document}

%% file: figures/teaser.tex
\twocolumn[{%
\renewcommand\twocolumn[1][]{#1}%
    \maketitle
    \newlength{\wid}
    \setlength{\wid}{0.106\textwidth}
    \addtolength{\tabcolsep}{-4pt}
    \begin{center}
        \vspace{-10pt}
        \centering
        \begin{tabular}{c:cccc:ccc}
            \includegraphics[align=c,width=\wid]{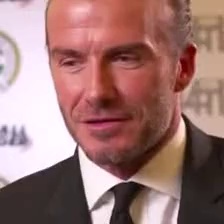}
            $\;$&$\;$
            \includegraphics[align=c,bmargin=0.13cm,width=\wid]{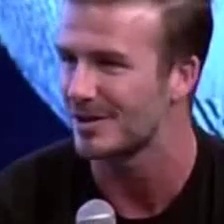} &
            \includegraphics[align=c,width=\wid]{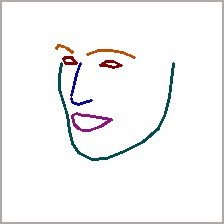}&
            \includegraphics[align=c,width=\wid]{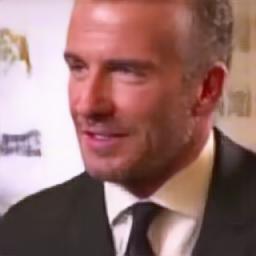}
            $\;\;$&$\;\;$
            \includegraphics[align=c,width=\wid]{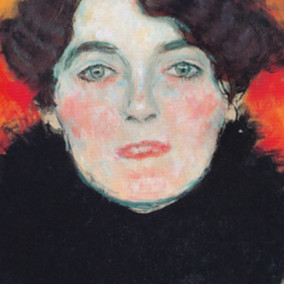}
            $\;$&$\;$
            \includegraphics[align=c,bmargin=0.13cm,width=\wid]{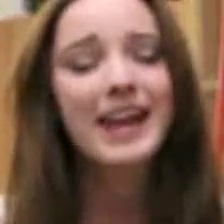} &
            \includegraphics[align=c,width=\wid]{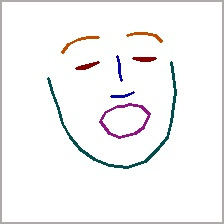} &
            \includegraphics[align=c,width=\wid]{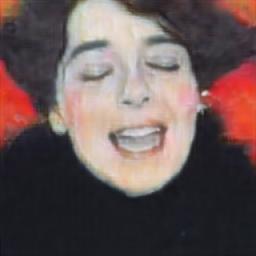}\\
            
            \includegraphics[align=c,width=\wid]{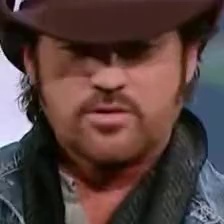}
            $\;$&$\;$
            \includegraphics[align=c,bmargin=0.13cm,width=\wid]{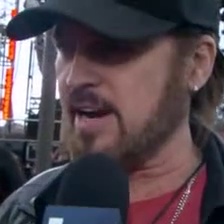} &
            \includegraphics[align=c,width=\wid]{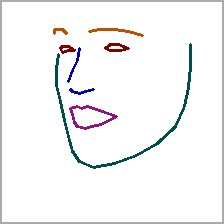}&
            \includegraphics[align=c,width=\wid]{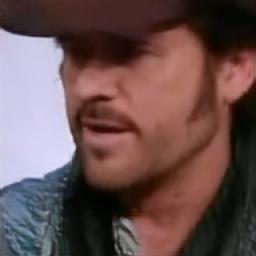}
            $\;\;$&$\;\;$
            \includegraphics[align=c,width=\wid]{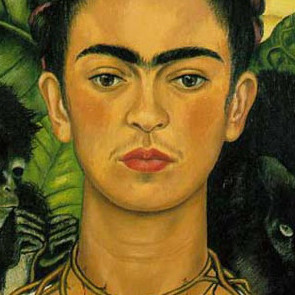}
            $\;$&$\;$
            \includegraphics[align=c,bmargin=0.13cm,width=\wid]{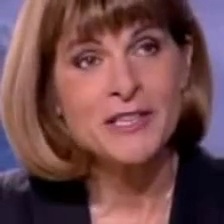} &
            \includegraphics[align=c,width=\wid]{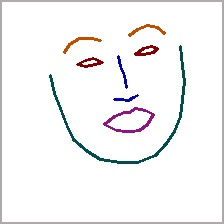}&
            \includegraphics[align=c,width=\wid]{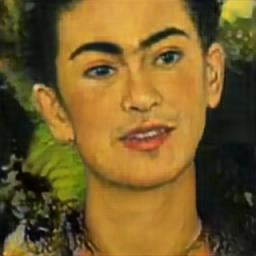}
            \vspace{1pt}\\
            \textbf{Source} $\;$& \multicolumn{3}{c}{\textbf{Target} $\,\rightarrow\,$ \textbf{Landmarks} $\,\rightarrow\,$ \textbf{Result}} &$\;$ \textbf{Source} $\;$& \multicolumn{3}{c}{\;\textbf{Target} $\,\rightarrow\,$ \textbf{Landmarks} $\,\rightarrow\,$ \textbf{Result}}
        \end{tabular}\vspace{-2pt}
        \captionof{figure}{The results of talking head image synthesis using face landmark tracks extracted from a different video sequence of the same person (on the left), and using face landmarks of a different person (on the right). The \textbf{results} are conditioned on the \textbf{landmarks} taken from the \textbf{target} frame, while the \textbf{source} frame is an example from the training set. The talking head models on the left were trained using eight frames, while the models on the right were trained in a one-shot manner.}
        \label{fig:teaser}
    \end{center}
}]

%% file: intro.tex
\section{Introduction}

In this work, we consider the task of creating personalized photorealistic talking head models, i.e.\ systems that can synthesize plausible video-sequences of speech expressions and mimics of a particular individual. More specifically, we consider the problem of synthesizing photorealistic personalized head images given a set of face landmarks, which drive the animation of the model. Such ability has practical applications for telepresence, including video-conferencing and multi-player games, as well as special effects industry. Synthesizing realistic talking head sequences is known to be hard for two reasons. First, human heads have high photometric, geometric and kinematic complexity. This complexity stems not only from modeling faces (for which a large number of modeling approaches exist) but also from modeling mouth cavity, hair, and garments. The second complicating factor is the acuteness of the human visual system towards even minor mistakes in the appearance modeling of human heads (the so-called \textit{uncanny valley} effect~\cite{Mori70}). Such low tolerance to modeling mistakes explains the current prevalence of non-photorealistic cartoon-like avatars in many practically-deployed teleconferencing systems.

To overcome the challenges, several works have proposed to synthesize articulated head sequences by warping a single or multiple static frames. Both classical warping algorithms~\cite{Averbuch17,Seitz96} and warping fields synthesized using machine learning (including deep learning)~\cite{Ganin16,Shu18,Wiles18} can be used for such purposes. While warping-based systems can create talking head sequences from as little as a single image, the amount of motion, head rotation, and disocclusion that they can handle without noticeable artifacts is limited.

\textit{Direct} (warping-free) synthesis of video frames using adversarially-trained deep convolutional networks (ConvNets) presents the new hope for photorealistic talking heads. Very recently, some remarkably realistic results have been demonstrated by such systems~\cite{Isola17,Kim18,Wang18c}. However, to succeed, such methods have to train large networks, where both generator and discriminator have tens of millions of parameters for each talking head. These systems, therefore, require a several-minutes-long video~\cite{Kim18,Wang18c} or a large dataset of photographs~\cite{Isola17} as well as hours of GPU training in order to create a new personalized talking head model. While this effort is lower than the one required by systems that construct photo-realistic head models using sophisticated physical and optical modeling~\cite{alexander2010digital}, it is still excessive for most practical telepresence scenarios, where we want to enable users to create their personalized head models with as little effort as possible. 

In this work, we present a system for creating talking head models from a handful of photographs (so-called \textit{few-shot learning}) and with limited training time. In fact, our system can generate a reasonable result based on a single photograph (\textit{one-shot learning}), while adding a few more photographs increases the fidelity of personalization. Similarly to \cite{Isola17,Kim18,Wang18c}, the talking heads created by our model are deep ConvNets that synthesize video frames in a direct manner by a sequence of convolutional operations rather than by warping. The talking heads created by our system can, therefore, handle a large variety of poses that goes beyond the abilities of warping-based systems.

The few-shot learning ability is obtained through extensive pre-training (\textit{meta-learning}) on a large corpus of talking head videos corresponding to different speakers with diverse appearance. In the course of meta-learning, our system simulates few-shot learning tasks and learns to transform landmark positions into realistically-looking personalized photographs, given a small training set of images with this person. After that, a handful of photographs of a new person sets up a new adversarial learning problem with high-capacity generator and discriminator pre-trained via meta-learning. The new adversarial problem converges to the state that generates realistic and personalized images after a few training steps.

In the experiments, we provide comparisons of talking heads created by our system with alternative neural talking head models \cite{Isola17,Wiles18} via quantitative measurements and a user study, where our approach generates images of sufficient realism and personalization fidelity to deceive the study participants. We demonstrate several uses of our talking head models, including video synthesis using landmark tracks extracted from video sequences of the same person, as well as \textit{puppeteering} (video synthesis of a certain person based on the face landmark tracks of a different person).

%% file: related.tex
\section{Related work}

A huge body of works is devoted to statistical modeling of the appearance of human faces~\cite{Blanz99}, with remarkably good results obtained both with classical techniques~\cite{Thies16} and, more recently, with deep learning~\cite{Lombardi18, Nagano18} (to name just a few). While modeling faces is a highly related task to talking head modeling, the two tasks are not identical, as the latter also involves modeling non-face parts such as hair, neck, mouth cavity and often shoulders/upper garment. These non-face parts cannot be handled by some trivial extension of the face modeling methods since they are much less amenable for registration and often have higher variability and higher complexity than the face part. In principle, the results of face modeling~\cite{Thies16} or lips modeling~\cite{Suwajanakorn17} can be stitched into an existing head video. Such design, however, does not allow full control over the head rotation in the resulting video and therefore does not result in a fully-fledged talking head system.

The design of our system borrows a lot from the recent progress in generative modeling of images. Thus, our architecture uses adversarial training~\cite{Goodfellow14} and, more specifically, the ideas behind conditional discriminators~\cite{Mirza14}, including projection discriminators~\cite{Miyato18a}. Our meta-learning stage uses the adaptive instance normalization mechanism~\cite{Huang17}, which was shown to be useful in large-scale conditional generation tasks~\cite{Brock18,Karras18b}. We also find an idea of content-style decomposition~\cite{Huang18} to be extremely useful for separating the texture from the body pose.

The model-agnostic meta-learner (MAML)~\cite{Finn17} uses meta-learning to obtain the initial state of an image classifier, from which it can quickly converge to image classifiers of unseen classes, given few training samples. This high-level idea is also utilized by our method, though our implementation of it is rather different. Several works have further proposed to combine adversarial training with meta-learning. Thus, data-augmentation GAN~\cite{Antoniou18}, MetaGAN~\cite{Zhang18a}, adversarial meta-learning~\cite{Yin18} use adversarially-trained networks to generate additional examples for classes unseen at the meta-learning stage. While these methods are focused on boosting the few-shot classification performance, our method deals with the training of image generation models using similar adversarial objectives. To summarize, we bring the adversarial fine-tuning into the meta-learning framework. The former is applied after we obtain initial state of the generator and the discriminator networks via the meta-learning stage.

Finally, very related to ours are the two recent works on text-to-speech generation~\cite{Arik18,Jia18}. Their setting (few-shot learning of generative models) and some of the components (standalone embedder network, generator fine-tuning) are are also used in our case. Our work differs in the application domain, the use of adversarial learning, its adaptation to the meta-learning process and implementation details.

%% file: method.tex
\begin{figure*}
    \centering
    \includegraphics[width=\textwidth]{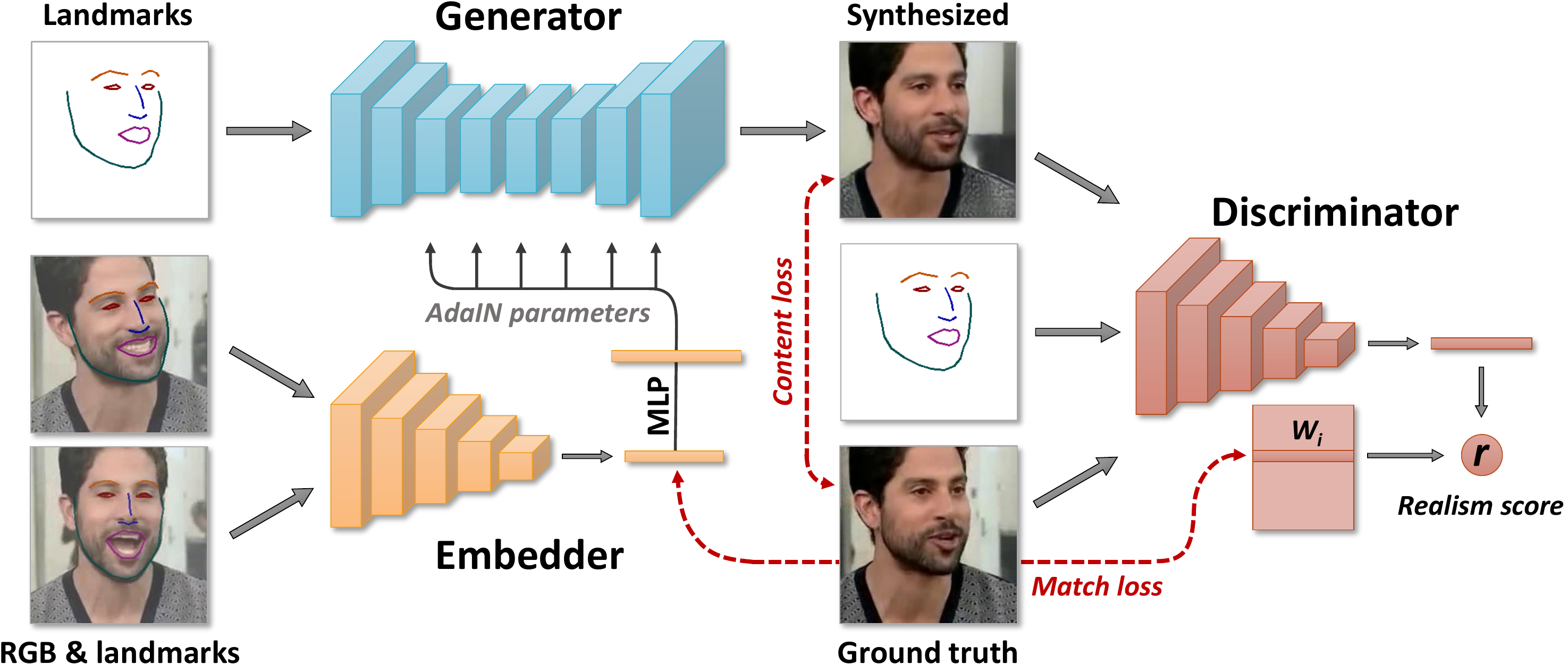}\vspace{-2pt}
    \caption{Our meta-learning architecture involves the embedder network that maps head images (with estimated face landmarks) to the embedding vectors, which contain pose-independent information. The generator network maps input face landmarks into output frames through the set of convolutional layers, which are modulated by the embedding vectors via adaptive instance normalization. During meta-learning, we pass sets of frames from the same video through the embedder, average the resulting embeddings and use them to predict adaptive parameters of the generator. Then, we pass the landmarks of a different frame through the generator, comparing the resulting image with the ground truth. Our objective function includes perceptual and adversarial losses, with the latter being implemented via a conditional projection discriminator.\vspace{-4pt}}
    \label{fig:metalearn}
\end{figure*}

\section{Methods}

\newcommand{\floor}[1]{\lfloor #1 \rfloor}
\newcommand{\ceil}[1]{\lceil #1 \rceil}
\renewcommand{\l}{}
\newcommand{\Strut}{\rule[-.4\baselineskip]{0pt}{\baselineskip}}

\def\x{\mathbf{x}}
\def\y{\mathbf{y}}
\def\e{\mathbf{e}}
\def\v{\mathbf{v}}
\def\w{\mathbf{w}}
\def\W{\mathbf{W}}
\def\P{\mathbf{P}}
\def\adv{\mathcal{L}_\text{ADV}}
\def\aadv{\mathcal{L'}_\text{ADV}}
\def\cnt{\mathcal{L}_\text{CNT}}
\def\mtch{\mathcal{L}_\text{MCH}}
\def\fm{\mathcal{L}_\text{FM}}
\def\disc{\mathcal{L}_\text{DSC}}
\def\new{_\text{NEW}}

\subsection{Architecture and notation}

The meta-learning stage of our approach assumes the availability of $M$ video sequences, containing talking heads of different people. We denote with $\x_i$ the $i$-th video sequence and with $\x_i(t)$ its $t$-th frame. During the learning process, as well as during test time, we assume the availability of the face landmarks' locations for all frames (we use an off-the-shelf face alignment code~\cite{Bulat17} to obtain them). The landmarks are rasterized into three-channel images using a predefined set of colors to connect certain landmarks with line segments. We denote with $\y_i(t)$ the resulting \textit{landmark image} computed for $\x_i(t)$.

In the meta-learning stage of our approach, the following three networks are trained (\fig{metalearn}):
\begin{itemize}[noitemsep,nolistsep,leftmargin=*]
    \item The \textit{embedder}  $E(\x_i(s),\y_i(s);\,\phi)$ takes a video frame $\x_i(s)$, an associated landmark image $\y_i(s)$ and maps these inputs into an $N$-dimensional vector $\hat{\e}_i(s)$. Here, $\phi$ denotes network parameters that are learned in the meta-learning stage. In general, during meta-learning, we aim to learn $\phi$ such that the vector $\hat{\e}_i(s)$ contains video-specific information (such as the person's identity) that is invariant to the pose and mimics in a particular frame $s$. We denote embedding vectors computed by the embedder as $\hat{\e}_i$.
    
    \item The \textit{generator}  $G(\y_i(t), \hat{\e}_i;\,\psi, \P)$ takes the landmark image $\y_i(t)$ for the video frame not seen by the embedder, the predicted video embedding $\hat{\e}_i$ and outputs a synthesized video frame $\hat{\x}_i(t)$. The generator is trained to maximize the similarity between its outputs and the ground truth frames. All parameters of the generator are split into two sets: the person-generic parameters $\psi$, and the person-specific parameters $\hat\psi_i$. During meta-learning, only $\psi$ are trained directly, while $\hat\psi_i$ are predicted from the embedding vector $\hat\e_i$ using a trainable projection matrix $\P$: $\hat\psi_i=\P\hat\e_i$.
    
    \item The \textit{discriminator} $D(\x_i(t),\y_i(t),i;\,\theta,\W,\w_0,b)$ takes a video frame $\x_i(t)$, an associated landmark image $\y_i(t)$ and the index of the training sequence $i$. Here, $\theta,\W,\w_0$ and $b$ denote the learnable parameters associated with the discriminator.  The discriminator contains a ConvNet part $V(\x_i(t),\y_i(t);\,\theta)$ that maps the input frame and the landmark image into an $N$-dimensional vector. The discriminator predicts a single scalar (realism score) $r$, that indicates whether the input frame $\x_i(t)$ is a real frame of the $i$-th video sequence and whether it matches the input pose $\y_i(t)$, based on the output of its ConvNet part and the parameters $\W,\w_0,b$.
\end{itemize}

\subsection{Meta-learning stage}

During the meta-learning stage of our approach, the parameters of all three networks are trained in an adversarial fashion. It is done by simulating episodes of $K$-shot learning ($K=8$ in our experiments). In each episode, we randomly draw a training video sequence $i$ and a single frame $t$ from that sequence. In addition to $t$, we randomly draw additional $K$ frames $s_1,s_2,\dots,s_K$ from the same sequence. We then compute the estimate $\hat{\e}_i$ of the $i$-th video embedding by simply averaging the embeddings $\hat{\e}_i(s_k)$ predicted for these additional frames:
\begin{equation} \label{eq:embed}
    \hat\e_i = \frac{1}{K} \sum_{k=1}^K E\left(\x_i(s_k),\y_i(s_k);\,\phi\right)\,.
\end{equation}

A reconstruction $\hat\x_i(t)$ of the $t$-th frame, based on the estimated embedding $\hat\e_i$, is then computed:
\begin{equation} \label{eq:embed2}
    \hat\x_i(t) = G\left(\y_i(t),\hat\e_i;\,\psi,\P\right)\,.
\end{equation}

The parameters of the embedder and the generator are then optimized to minimize the following objective that comprises the content term, the adversarial term, and the embedding match term:
\begin{align} \label{eq:totalloss}
    \mathcal{L}(\phi,\psi,&\P,\theta,\W,\w_0,b) = \cnt(\phi,\psi,\P) + \\ &\adv(\phi,\psi,\P,\theta,\W,\w_0,b)+\mtch(\phi, \W)\,. \nonumber
\end{align}

In \eq{totalloss}, the content loss term $\cnt$ measures the distance between the ground truth image $\x_i(t)$ and the reconstruction $\hat\x_i(t)$ using the perceptual similarity measure~\cite{Johnson16}, corresponding to VGG19~\cite{Simonyan15} network trained for ILSVRC classification and VGGFace~\cite{Parkhi15} network trained for face verification. The loss is calculated as the weighted sum of $L_1$ losses between the features of these networks.

The adversarial term in \eq{totalloss} corresponds to the realism score computed by the discriminator, which needs to be maximized, and a feature matching term~\cite{Wang18b}, which essentially is a perceptual similarity measure, computed using discriminator (it helps with the stability of the training):
\begin{align} \label{eq:advers}
    \adv(\phi,\psi,\P,\theta,\W,\w_0,b) = &\\ -D(\hat\x_i(t),\y_i(t),i;&\,\theta,\W,\w_0,b) + \fm\,. \nonumber
\end{align}
Following the projection discriminator idea \cite{Miyato18a}, the columns of the matrix $\W$ contain the embeddings that correspond to individual videos. The discriminator first maps its inputs to an $N$-dimensional vector $V(\x_i(t),\y_i(t);\theta)$ and then computes the realism score as: 
\begin{align} \label{eq:metadisc}
    D(\hat\x_i(t),\y_i(t),i;\,\theta,\W,\w_0,b) =&\\ V\left(\hat\x_i(t),\y_i(t);\,\theta\right)^T&(\W_i+\w_0)+b\,,\nonumber
\end{align}
where $\W_i$ denotes the $i$-th column of the matrix $\W$. At the same time, $\w_0$ and $b$ do not depend on the video index, so these terms correspond to the general realism of $\hat\x_i(t)$ and its compatibility with the landmark image $\y_i(t)$.

Thus, there are two kinds of video embeddings in our system: the ones computed by the embedder, and the ones that correspond to the columns of the matrix $\W$ in the discriminator. The match term $\mtch(\phi, \W)$ in \eq{totalloss} encourages the similarity of the two types of embeddings by penalizing the $L_1$-difference between $E\left(\x_i(s_k),\y_i(s_k);\,\phi\right)$ and $\W_i$.

As we update the parameters $\phi$ of the embedder and the parameters $\psi$ of the generator, we also update the parameters $\theta,\W,\w_0,b$ of the discriminator. The update is driven by the minimization of the following hinge loss, which encourages the increase of the realism score on real images $\x_i(t)$ and its decrease on synthesized images $\hat\x_i(t)$:
\begin{align} \label{eq:metahinge}
   \disc&(\phi,\psi,\P,\theta,\W,\w_0,b) = \\
   &\max(0,1+D(\hat\x_i(t),\y_i(t),i;\,\phi,\psi,\theta,\W,\w_0,b)) +\nonumber\\
   &\qquad\,\max(0,1-D(\x_i(t),\y_i(t),i;\,\theta,\W,\w_0,b))\,.\nonumber
\end{align}
The objective \eq{metahinge} thus compares the realism of the fake example $\hat\x_i(t)$ and the real example $\x_i(t)$ and then updates the discriminator parameters to push these scores below $-1$ and above $+1$ respectively. The training proceeds by alternating updates of the embedder and the generator that minimize the losses $\cnt,\adv$ and $\mtch$ with the updates of the discriminator that minimize the loss $\disc$.

\subsection{Few-shot learning by fine-tuning}

Once the meta-learning has converged, our system can learn to synthesize talking head sequences for a new person, unseen during meta-learning stage. As before, the synthesis is conditioned on the landmark images. The system is learned in a few-shot way, assuming that $T$ training images $\x(1),\x(2),\dots,\x(T)$ (e.g.\ $T$ frames of the same video) are given and that $\y(1),\y(2),\dots,\y(T)$ are the corresponding landmark images. Note that the number of frames $T$ needs not to be equal to $K$ used in the meta-learning stage. 

Naturally, we can use the meta-learned embedder to estimate the embedding for the new talking head sequence:
\begin{equation} \label{eq:init}
    \hat\e\new = \frac{1}{T} \sum_{t=1}^T E(\x(t),\y(t);\,\phi)\,,
\end{equation}
reusing the parameters $\phi$ estimated in the meta-learning stage. A straightforward way to generate new frames, corresponding to new landmark images, is then to apply the generator using the estimated embedding $\hat\e\new$ and the meta-learned parameters $\psi$, as well as projection matrix $\P$. By doing so, we have found out that the generated images are plausible and realistic, however, there often is a considerable identity gap that is not acceptable for most applications aiming for high personalization degree. 

This identity gap can often be bridged via the \textit{fine-tuning stage}. The fine-tuning process can be seen as a simplified version of meta-learning with a single video sequence and a smaller number of frames. The fine-tuning process involves the following components:
\begin{itemize}[noitemsep,nolistsep,leftmargin=*]
    \item The generator $G(\y(t),\hat\e\new;\,\psi,\P)$ is now replaced with $G'(\y(t);\,\psi,\psi')$. As before, it takes the landmark image $\y(t)$ and outputs the synthesized frame $\hat\x(t)$. Importantly, the person-specific generator parameters, which we now denote with $\psi'$, are now directly optimized alongside the person-generic parameters $\psi$. We still use the computed embeddings $\hat\e\new$ and the projection matrix $\P$ estimated at the meta-learning stage to initialize $\psi'$, i.e.\ we start with $\psi'=\P\hat\e\new$.
     
     \item The discriminator $D'(\x(t),\y(t);\theta,\w',b)$, as before, computes the realism score. Parameters $\theta$ of its ConvNet part $V(\x(t),\y(t);\theta)$ and bias $b$ are initialized to the result of the meta-learning stage. The initialization of $\w'$ is discussed below.
\end{itemize}
During fine-tuning, the realism score of the discriminator is obtained in a similar way to the meta-learning stage:
\begin{align} \label{eq:ftdisc}
    D'(\hat\x(t),\y(t);\,\theta,\w',b) =&\\ V\left(\hat\x(t),\y(t);\,\theta\right)^T&\w'+b\,.\nonumber
\end{align}
As can be seen from the comparison of expressions \eq{metadisc} and \eq{ftdisc}, the role of the vector $\w'$ in the fine-tuning stage is the same as the role of the vector $\W_i+\w_0$ in the meta-learning stage. For the intiailization, we do not have access to the analog of $\W_i$ for the new personality (since this person is not in the meta-learning dataset). However, the  match term $\mtch{}$ in the meta-learning process ensures the similarity between the discriminator video-embeddings and the vectors computed by the embedder. Hence, we can initialize $\w'$ to the sum of $\w_0$ and $\hat\e\new$.

Once the new learning problem is set up, the loss functions of the fine-tuning stage follow directly from the meta-learning variants. Thus, the generator parameters $\psi$ and $\psi'$ are optimized to minimize the simplified objective:
\begin{align} \label{eq:finetune}
    \mathcal{L}'(\psi,\psi'&,\theta,\w',b) = \\ &\cnt'(\psi,\psi') +
    \aadv{}(\psi,\psi',\theta,\w',b) \,,\nonumber
\end{align}
where $t \in \{1\dots T\}$ is the number of the training example. The discriminator parameters $\theta,\w\new,b$ are optimized by minimizing the same hinge loss as in \eq{metahinge}:
\begin{align}
   &\mathcal{L}'_\text{DSC}(\psi,\psi',\theta,\w',b) =\\     &\max(0,1+D(\hat\x(t),\y(t);\,\psi,\psi',\theta,\w',b)) +\nonumber\\
   &\qquad\qquad\!\max(0,1-D(\x(t),\y(t);\,\theta,\w',b))\,.\nonumber
\end{align}

In most situations, the fine-tuned generator provides a much better fit of the training sequence. The initialization of all parameters via the meta-learning stage is also crucial. As we show in the experiments, such initialization injects a strong realistic talking head prior, which allows our model to extrapolate and predict realistic images for poses with varying head poses and facial expressions.

\subsection{Implementation details}

We base our generator network $G(\y_i(t), \hat\e_i; \psi, \P)$ on the image-to-image translation architecture proposed by Johnson et.\ al.~\cite{Johnson16}, but replace downsampling and upsampling layers with residual blocks similarly to~\cite{Brock18} (with batch normalization~\cite{Ioffe15} replaced by instance normalization~\cite{Ulyanov16}). The person-specific parameters $\hat\psi_i$ serve as the affine coefficients of instance normalization layers, following the adaptive instance normalization technique proposed in~\cite{Huang17}, though we still use regular (non-adaptive) instance normalization layers in the downsampling blocks that encode landmark images $\y_i(t)$.

For the embedder $E(\x_i(s), \y_i(s); \phi)$ and the convolutional part of the discriminator $V(\x_i(t), \y_i(t); \theta)$, we use similar networks, which consist of residual downsampling blocks (same as the ones used in the generator, but without normalization layers). The discriminator network, compared to the embedder, has an additional residual block at the end, which operates at $4\times4$ spatial resolution. To obtain the vectorized outputs in both networks, we perform global sum pooling over spatial dimensions followed by ReLU. 

We use spectral normalization~\cite{Miyato18b} for all convolutional and fully connected layers in all the networks. We also use self-attention blocks, following~\cite{Brock18} and~\cite{Zhang18b}. They are inserted at $32 \times 32$ spatial resolution in all downsampling parts of the networks and at $64 \times 64$ resolution in the upsampling part of the generator.

For the calculation of $\cnt$, we evaluate $L_1$ loss between activations of \texttt{Conv1,6,11,20,29} VGG19 layers and \texttt{Conv1,6,11,18,25} VGGFace layers for real and fake images. We sum these losses with the weights equal to $1.5\cdot10^{-1}$ for VGG19 and $2.5\cdot10^{-2}$ for VGGFace terms. We use Caffe~\cite{Jia14} trained versions for both of these networks. For $\fm$, we use activations after each residual block of the discriminator network and the weights equal to $10$. Finally, for $\mtch{}$ we also set the weight to $10$.

We set the minimum number of channels in convolutional layers to $64$ and the maximum number of channels as well as the size $N$ of the embedding vectors to $512$. In total, the embedder has 15 million parameters, the generator has 38 million parameters. The convolutional part of the discriminator has 20 million parameters. The networks are optimized using Adam~\cite{Diederik14}. We set the learning rate of the embedder and the generator networks to $5\times10^{-5}$ and to $2\times10^{-4}$ for the discriminator, doing two update steps for the latter per one of the former, following~\cite{Zhang18b}.


%% file: experiments.tex
\section{Experiments}

\begin{table}
    \begin{subtable}{\linewidth}
        \centering
        \begin{tabular}{ l c c c c}
            Method (T) & FID$\downarrow$ & SSIM$\uparrow$ & CSIM$\uparrow$ & USER$\downarrow$ \\
            \hline
            \multicolumn{5}{c}{VoxCeleb1}\\
            \hline
            X2Face (1)     & 45.8 & \textbf{0.68} & \textbf{0.16} & 0.82 \\
            Pix2pixHD (1)  & \textbf{42.7} & 0.56 & 0.09 & 0.82\\
            Ours  (1)      & 43.0 & 0.67 & 0.15 & \textbf{0.62}\\
            \hline
            X2Face (8)      & 51.5 & \textbf{0.73} & \textbf{0.17} & 0.83\\
            Pix2pixHD (8)   & \textbf{35.1} & 0.64 & 0.12 & 0.79\\
            Ours (8)        & 38.0 & 0.71 & \textbf{0.17} & \textbf{0.62}\\
            \hline
            X2Face (32)    & 56.5 & \textbf{0.75} & 0.18 & 0.85 \\
            Pix2pixHD (32) & \textbf{24.0} & 0.70 & 0.16 & 0.71 \\
            Ours (32)      & 29.5 & 0.74 & \textbf{0.19} & \textbf{0.61} \\
            \hline
            \multicolumn{5}{c}{VoxCeleb2} \\                                
            \hline
            Ours-FF (1)  & \textbf{46.1} & 0.61 & \textbf{0.42} & \textbf{0.43} \\
            Ours-FT (1)  & 48.5 & \textbf{0.64} & 0.35 & 0.46 \\
            \hline
            Ours-FF (8)  & 42.2 & 0.64 & \textbf{0.47} & 0.40 \\
            Ours-FT (8)  & \textbf{42.2} & \textbf{0.68} & 0.42 & \textbf{0.39} \\
            \hline
            Ours-FF (32) & 40.4 & 0.65 & \textbf{0.48} & 0.38 \\
            Ours-FT (32) & \textbf{30.6} & \textbf{0.72} & 0.45 & {\color{red}\textbf{0.33}} \\

        \end{tabular}
    \end{subtable}\vspace{-2pt}
    \caption{Quantitative comparison of methods on different datasets with multiple few-shot learning settings. Please refer to the text for more details and discussion.\vspace{-4pt}}\label{tab:maincomp}
\end{table}

Two datasets with talking head videos are used for quantitative and qualitative evaluation: VoxCeleb1~\cite{Nagrani17} (256p videos at 1 fps) and VoxCeleb2~\cite{Chung18} (224p videos at 25 fps), with the latter having approximately 10 times more videos than the former. VoxCeleb1 is used for comparison with baselines and ablation studies, while by using VoxCeleb2 we show the full potential of our approach. 

\paragraph{Metrics.}

For the quantitative comparisons, we fine-tune all models on few-shot learning sets of size $T$ for a person not seen during meta-learning (or pretraining) stage. After the few-shot learning, the evaluation is performed on the hold-out part of the same sequence (so-called \textit{self-reenactment} scenario). For the evaluation, we uniformly sampled 50 videos from VoxCeleb test sets and 32 hold-out frames for each of these videos (the fine-tuning and the hold-out parts do not overlap).

We use multiple comparison metrics to evaluate photo-realism and identity preservation of generated images. Namely, we use Frechet-inception distance (FID)~\cite{Guyon17}, mostly measuring perceptual realism, structured similarity (SSIM)~\cite{Wang04}, measuring low-level similarity to the ground truth images, and cosine similarity (CSIM) between embedding vectors of the state-of-the-art face recognition network~\cite{Deng19} for measuring identity mismatch (note that this network has quite different architecture from VGGFace used within content loss calculation during training).

We also perform a user study in order to evaluate perceptual similarity and realism of the results as seen by the human respondents. We show people the triplets of images of the same person taken from three different video sequences. Two of these images are real and one is fake, produced by one of the methods, which are being compared. We ask the user to find the fake image given that all of these images are of the same person. This evaluates both photo-realism and identity preservation because the user can infer the identity from the two real images (and spot an identity mismatch even if the generated image is perfectly realistic). We use the user accuracy (success rate) as our metric. The lower bound here is the accuracy of one third (when users cannot spot fakes based on non-realism or identity mismatch and have to guess randomly). Generally, we believe that this user-driven metric (USER) provides a much better idea of the quality of the methods compared to FID, SSIM, or CSIM.

\paragraph{Methods.}

On the VoxCeleb1 dataset we compare our model against two other systems: X2Face~\cite{Wiles18} and Pix2pixHD~\cite{Wang18b}. For X2Face, we have used the model, as well as pretrained weights, provided by the authors (in the original paper it was also trained and evaluated on the VoxCeleb1 dataset). For Pix2pixHD, we pretrained the model from scratch on the whole dataset for the same amount of iterations as our system without any changes to thetraining pipeline proposed by the authors. We picked X2Face as a strong baseline for warping-based methods and Pix2pixHD for direct synthesis methods.

In our comparison, we evaluate the models in several scenarios by varying the number of frames $T$ used in few-shot learning. X2Face, as a feed-forward method, is simply initialized via the training frames, while Pix2pixHD and our model are being additionally fine-tuned for 40 epochs on the few-shot set. Notably, in the comparison, X2Face uses dense correspondence field, computed on the ground truth image, to synthesize the generated one, while our method and Pix2pixHD use very sparse landmark information, which arguably gives X2Face an unfair advantage.

\paragraph{Comparison results.}

\begin{figure*}
    \centering    
    \setlength{\wid}{0.179\textwidth}
    \addtolength{\tabcolsep}{-4pt}
    \begin{tabular}{m{0.6cm}c:cccc}
        \centering{\textbf{1}}&
        \includegraphics[align=c,bmargin=0.13cm,width=\wid]{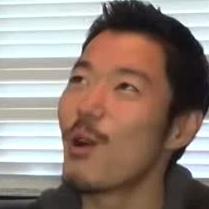}
        \;&\;
        \includegraphics[align=c,width=\wid]{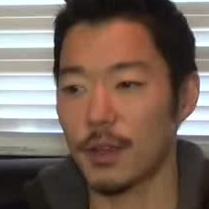}&
        \includegraphics[align=c,width=\wid]{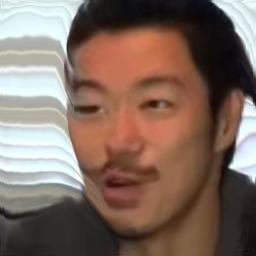}&
        \includegraphics[align=c,width=\wid]{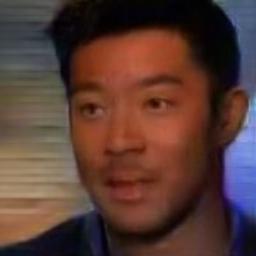}&
        \includegraphics[align=c,width=\wid]{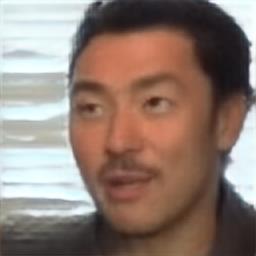}\\
        \centering{\textbf{8}}&
        \includegraphics[align=c,bmargin=0.13cm,width=\wid]{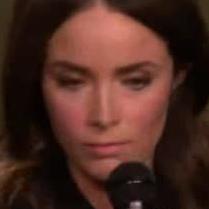}
        \;&\;
        \includegraphics[align=c,width=\wid]{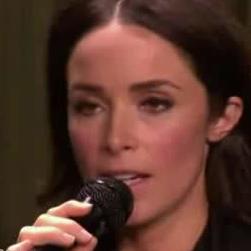}&
        \includegraphics[align=c,width=\wid]{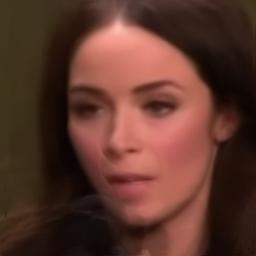}&
        \includegraphics[align=c,width=\wid]{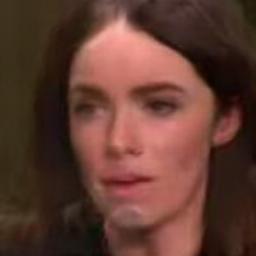}&
        \includegraphics[align=c,width=\wid]{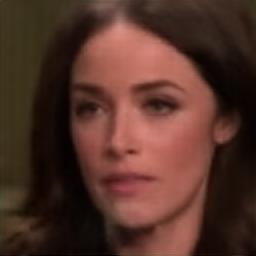}\\
        \centering{\textbf{32}}&
        \includegraphics[align=c,bmargin=0.13cm,width=\wid]{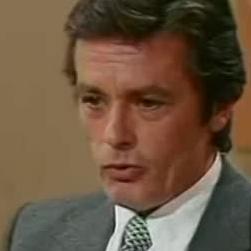}
        \;&\;
        \includegraphics[align=c,width=\wid]{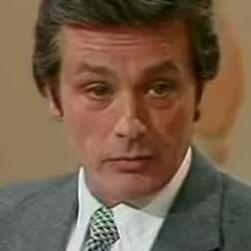}&
        \includegraphics[align=c,width=\wid]{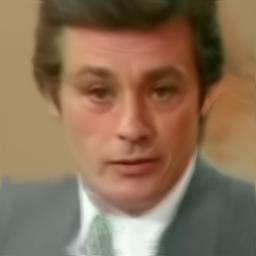}&
        \includegraphics[align=c,width=\wid]{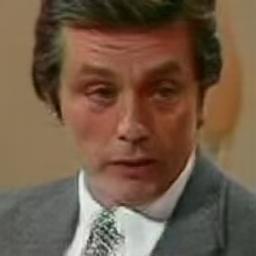}&
        \includegraphics[align=c,width=\wid]{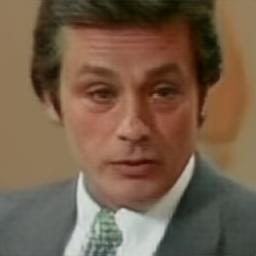}\\
        \centering{\textbf{T}} & \textbf{Source} \;&\; \textbf{Ground truth} & \textbf{X2Face} & \textbf{Pix2pixHD} & \textbf{Ours}
    \end{tabular}\vspace{-2pt}
    \caption{Comparison on the VoxCeleb1 dataset. For each of the compared methods, we perform one- and few-shot learning on a video of a person not seen during meta-learning or pretraining. We set the number of training frames equal to \textbf{T} (the leftmost column). One of the training frames is shown in the \textbf{source} column. Next columns show \textbf{ground truth} image, taken from the test part of the video sequence, and the generated results of the compared methods.\vspace{-4pt}}
    \label{fig:voxceleb1}
\end{figure*}

We perform comparison with baselines in three different setups, with 1, 8 and 32 frames in the fine-tuning set. Test set, as mentioned before, consists of 32 hold-out frames for each of the 50 test video sequences. Moreover, for each test frame we sample two frames at random from the other video sequences with the same person. These frames are used in triplets alongside with fake frames for user-study.

As we can see in \tab{maincomp}-Top, baselines consistently outperform our method on the two of our similarity metrics. We argue that this is intrinsic to the methods themselves: X2Face uses $L_2$ loss during optimization~\cite{Wiles18}, which leads to a good SSIM score. On the other hand, Pix2pixHD maximizes only perceptual metric, without identity preservation loss, leading to minimization of FID, but has bigger identity mismatch, as seen from the CSIM column. Moreover, these metrics do not correlate well with human perception, since both of these methods produce uncanny valley artifacts, as can be seen from qualitative comparison \fig{voxceleb1} and the user study results. Cosine similarity, on the other hand, better correlates with visual quality, but still favours blurry, less realistic images, and that can also be seen by comparing \tab{maincomp}-Top with the results presented in \fig{voxceleb1}.

While the comparison in terms of the objective metrics is inconclusive, the user study (that included 4800 triplets, each shown to 5 users) clearly reveals the much higher realism and personalization degree achieved by our method.

We have also carried out the ablation study of our system and the comparison of the few-shot learning timings. Both are provided in the Supplementary material.

\paragraph{Large-scale results.}

\begin{figure*}
    \centering    
    \setlength{\wid}{0.179\textwidth}
    \addtolength{\tabcolsep}{-4pt}
    \begin{tabular}{m{0.6cm}c:cccc}
        \centering{\textbf{1}}&
        \includegraphics[align=c,bmargin=0.13cm,width=\wid]{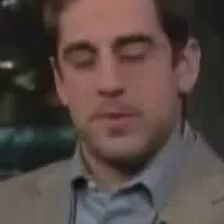}
        \;&\;
        \includegraphics[align=c,width=\wid]{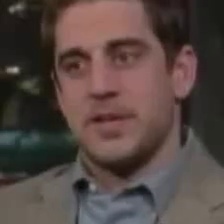}&
        \includegraphics[align=c,width=\wid]{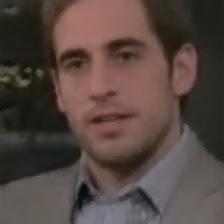}&
        \includegraphics[align=c,width=\wid]{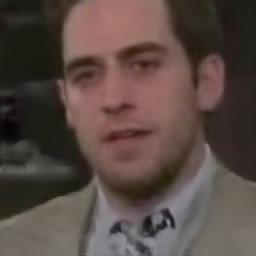}&
        \includegraphics[align=c,width=\wid]{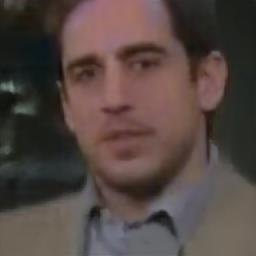}\\
        \centering{\textbf{8}}&
        \includegraphics[align=c,bmargin=0.13cm,width=\wid]{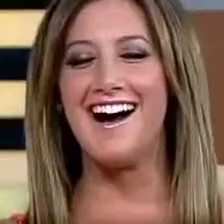}
        \;&\;
        \includegraphics[align=c,width=\wid]{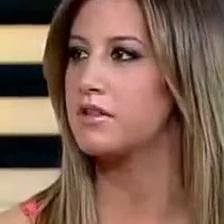}&
        \includegraphics[align=c,width=\wid]{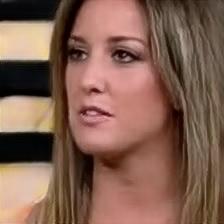}&
        \includegraphics[align=c,width=\wid]{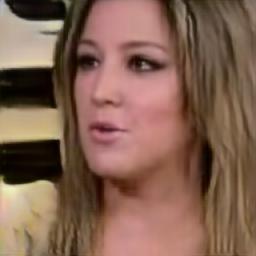}&
        \includegraphics[align=c,width=\wid]{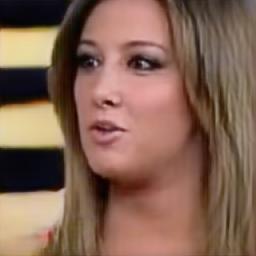}\\
        \centering{\textbf{32}}&
        \includegraphics[align=c,bmargin=0.13cm,width=\wid]{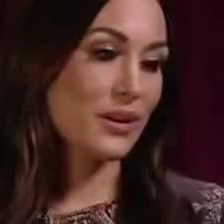}
        \;&\;
        \includegraphics[align=c,width=\wid]{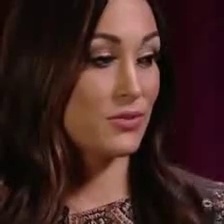}&
        \includegraphics[align=c,width=\wid]{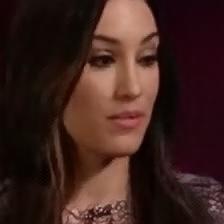}&
        \includegraphics[align=c,width=\wid]{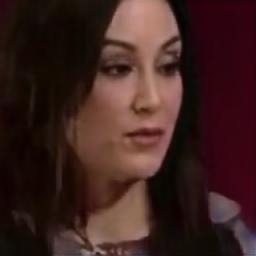}&
        \includegraphics[align=c,width=\wid]{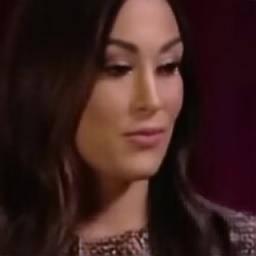}\\
        \centering{\textbf{T}} & \textbf{Source} \;&\; \textbf{Ground truth} & \textbf{Ours-FF} & \textbf{\begin{tabular}{c}Ours-FT\\ before fine-tuning\end{tabular}} & \textbf{\begin{tabular}{c}Ours-FT\\ after fine-tuning\end{tabular}}
    \end{tabular}\vspace{-2pt}
    \caption{Results for our best models on the VoxCeleb2 dataset. The number of training frames is, again, equal to \textbf{T} (the leftmost column) and the example training frame in shown in the \textbf{source} column. Next columns show \textbf{ground truth} image and the results for \textbf{Ours-FF} feed-forward model, \textbf{Ours-FT} model \textbf{before} and \textbf{after fine-tuning}. While the feed-forward variant allows fast (real-time) few-shot learning of new avatars, fine-tuning ultimately provides better realism and fidelity.\vspace{-4pt}}
    \label{fig:voxceleb2}
\end{figure*}

We then scale up the available data and train our method on a larger VoxCeleb2 dataset. Here, we train two variants of our method. FF (feed-forward) variant is trained for 150 epochs without the embedding matching loss $\mtch$ and, therefore, we only use it without fine-tuning (by simply predicting adaptive parameters $\psi'$ via the projection of the embedding $\hat\e\new$). The FT variant is trained for half as much (75 epochs) but with $\mtch$, which allows fine-tuning. We run the evaluation for both of these models since they allow to trade off few-shot learning speed versus the results quality. Both of them achieve considerably higher scores, compared to smaller-scale models trained on VoxCeleb1. Notably, the FT model reaches the lower bound of $0.33$ for the user study accuracy in $T=32$ setting, which is a perfect score. We present results for both of these models in \fig{voxceleb2} and more results (including results, where animation is driven by landmarks from a different video of the same person) are given in the supplementary material and in \fig{teaser}.

Generally, judging by the results of comparisons (\tab{maincomp}-Bottom) and the visual assessment, the FF model performs better for low-shot learning (e.g.\ one-shot), while the FT model achieves higher quality for bigger $T$ via adversarial fine-tuning.

\paragraph{Puppeteering results.}

\begin{figure}
    \centering
    \newcommand\wi{0.106}
    \begin{tabular}{c:ccc}
        \includegraphics[width=\wi\textwidth]{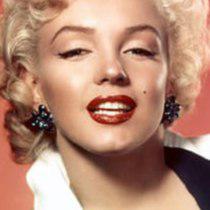}
        \;&\;
        \includegraphics[width=\wi\textwidth]{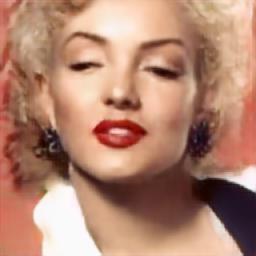}&
        \includegraphics[width=\wi\textwidth]{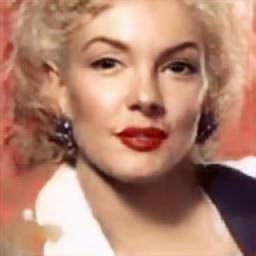}&
        \includegraphics[width=\wi\textwidth]{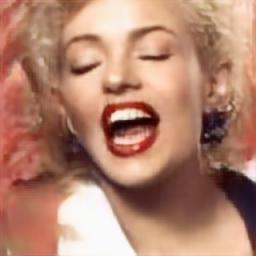}\\
        \includegraphics[width=\wi\textwidth]{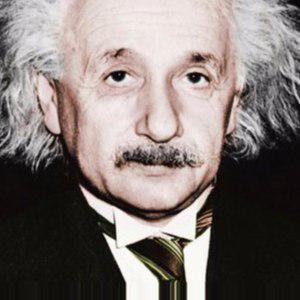}
        \;&\;
        \includegraphics[width=\wi\textwidth]{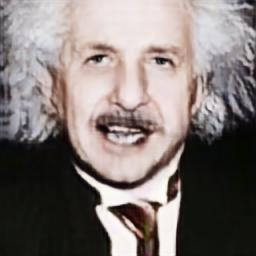}&
        \includegraphics[width=\wi\textwidth]{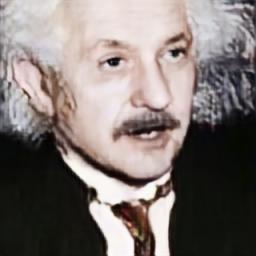}&
        \includegraphics[width=\wi\textwidth]{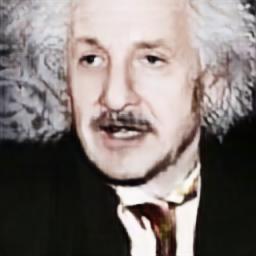}\\
        \includegraphics[width=\wi\textwidth]{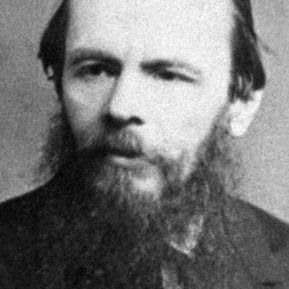}
        \;&\;
        \includegraphics[width=\wi\textwidth]{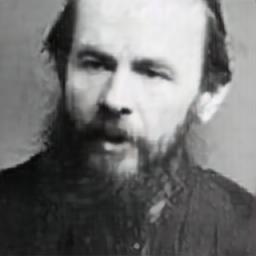}&
        \includegraphics[width=\wi\textwidth]{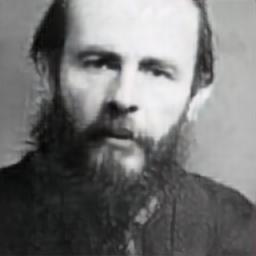}&
        \includegraphics[width=\wi\textwidth]{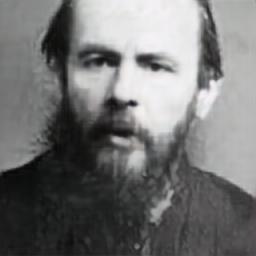}\\  
        \textbf{Source} & \multicolumn{3}{c}{\textbf{Generated images}}
    \end{tabular}\vspace{-2pt}
    \caption{Bringing still photographs to life. We show the puppeteering results for one-shot models learned from photographs in the \textbf{source} column. Driving poses were taken from the VoxCeleb2 dataset. Digital zoom recommended.\vspace{-4pt}}
    \label{fig:livingportraits}
\end{figure}

Finally, we show the results for the puppeteering of photographs and paintings. For that, we evaluate the model, trained in one-shot setting, on poses from test videos of the VoxCeleb2 dataset. We rank these videos using CSIM metric, calculated between the original image and the generated one. This allows us to find persons with similar geometry of the landmarks and use them for the puppeteering. The results can be seen in \fig{livingportraits} as well as in \fig{teaser}.

\addtolength{\tabcolsep}{-4pt}
\addtolength{\tabcolsep}{4pt}

%% file: conclusion.tex
\section{Conclusion}

We have presented a framework for meta-learning of adversarial generative models, which is able to train highly-realistic virtual talking heads in the form of deep generator networks. Crucially, only a handful of photographs (as little as one) is needed to create a new model, whereas the model trained on 32 images achieves perfect realism and personalization score in our user study (for 224p static images).

Currently, the key limitations of our method are the mimics representation (in particular, the current set of landmarks does not represent the gaze in any way) and the lack of landmark adaptation. Using landmarks from a different person leads to a noticeable personality mismatch. So, if one wants to create ``fake'' puppeteering videos without such mismatch, some landmark adaptation is needed. We note, however, that many applications do not require puppeteering a different person and instead only need the ability to drive one's own talking head. For such scenario, our approach already provides a high-realism solution.

\vfill

%% file: suppmat.tex
\clearpage

\section{Supplementary material}

In the supplementary material, we provide additional qualitative results as well as an ablation study and a time comparison between our method and the baselines for both inference and training.

\subsection{Time comparison results.}

In \tab{timecomp}, we provide a comparison of timings for the three methods. Additionally, we included the feed-forward variant of our method in the comparison, which was trained only for the VoxCeleb2 dataset. The comparison was carried out on a single NVIDIA P40 GPU. For Pix2pixHD and our method, few-shot learning was done via fine-tuning for 40 epochs on the training set of size $T$. For $T$ larger than 1, we trained the models on batches of $8$ images. Each measurement was averaged over 100 iterations.

We see that, given enough training data, our method in feed-forward variant can outpace all other methods by a large margin in terms of few-shot training time, while keeping personalization fidelity and realism of the outputs on quite a high level (as can be seen in \fig{voxceleb2}). But in order to achieve the best results in terms of quality, fine-tuning has to be performed, which takes approximately four and a half minutes on the P40 GPU for 32 training images. The number of epochs and, hence, the fine-tuning speed can be optimized further on a case by case basis or via the introduction of a training scheduler, which we did not perform.

On the other hand, inference speed for our method is comparable or slower than other methods, which is caused by a large number of parameters we need to encode the prior knowledge about talking heads. Though, this figure can be drastically improved via the usage of more modern GPUs (on an NVIDIA 2080 Ti, the inference time can be decreased down to 13ms per frame, which is enough for most real-time applications).

\begin{table}
    \begin{subtable}{\linewidth}
        \centering
        \newcolumntype{R}{>{\raggedleft\arraybackslash}X}
        \begin{tabularx}{\linewidth}{X R}
            Method (T) & Time, s \\
            \hline
            \multicolumn{2}{c}{Few-shot learning} \\
            \hline
            X2Face     (1) & 0.236 \\
            Pix2pixHD  (1) & 33.92 \\
            Ours       (1) & 43.84 \\
            Ours-FF    (1) & \textbf{0.061} \\
            \hline
            X2Face     (8) & 1.176 \\
            Pix2pixHD  (8) & 52.40 \\
            Ours       (8) & 85.48 \\
            Ours-FF    (8) & \textbf{0.138} \\
            \hline
            X2Face    (32) & 7.542 \\
            Pix2pixHD (32) & 122.6 \\
            Ours      (32) & 258.0 \\
            Ours-FF   (32) & \textbf{0.221} \\
            \hline
            \multicolumn{2}{c}{Inference} \\                                
            \hline
            X2Face         & 0.110 \\
            Pix2pixHD      & \textbf{0.034} \\
            Ours           & 0.139
        \end{tabularx}
    \end{subtable}
    \caption{Quantitative comparison of few-shot learning and inference timings for the three models.}\label{tab:timecomp}
    \vspace{-4pt}
\end{table}

\begin{figure*}[!h]
    \centering
    \setlength{\wid}{0.145\textwidth}
    \addtolength{\tabcolsep}{-4pt}
    \begin{tabular}{m{0.3cm} c : ccccc}
        \centering\rotatebox{90}{\textbf{Source}}&
        \includegraphics[align=c,bmargin=0.13cm,width=\wid,trim={0 0 0 147px},clip]{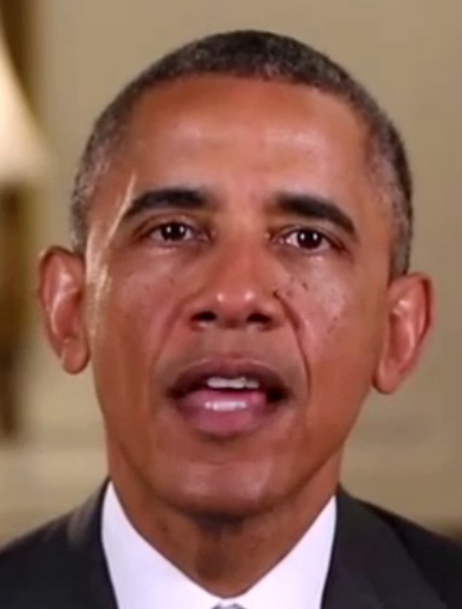}
        \;&\;
        \includegraphics[align=c,width=\wid,trim={20px 22px 20px 18px},clip]{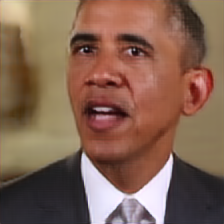}&
        \includegraphics[align=c,width=\wid,trim={20px 22px 20px 18px},clip]{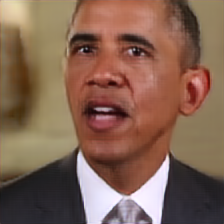}&
        \includegraphics[align=c,width=\wid,trim={20px 22px 20px 18px},clip]{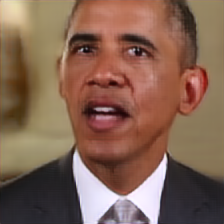}&
        \includegraphics[align=c,width=\wid,trim={20px 22px 20px 18px},clip]{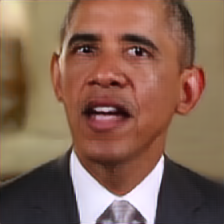}&
        \includegraphics[align=c,width=\wid,trim={20px 22px 20px 18px},clip]{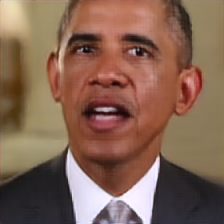}
        \\
        \centering\rotatebox{90}{\textbf{Face2Face}}&
        \includegraphics[align=c,bmargin=0.13cm,width=\wid,trim={0 0 0 147px},clip]{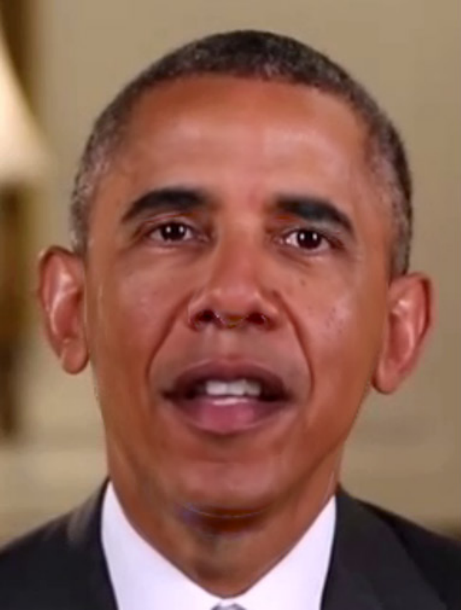}
        \;&\;
        \includegraphics[align=c,width=\wid,trim={20px 22px 20px 18px},clip]{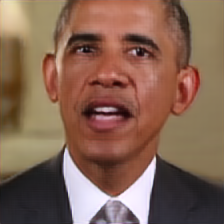}&
        \includegraphics[align=c,width=\wid,trim={20px 22px 20px 18px},clip]{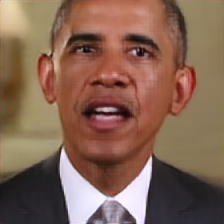}&
        \includegraphics[align=c,width=\wid,trim={20px 22px 20px 18px},clip]{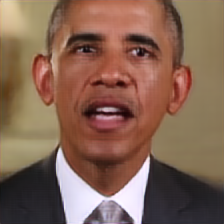}&
        \includegraphics[align=c,width=\wid,trim={20px 22px 20px 18px},clip]{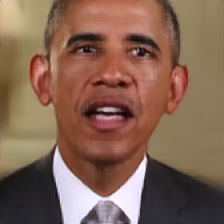}&
        \includegraphics[align=c,width=\wid,trim={20px 22px 20px 18px},clip]{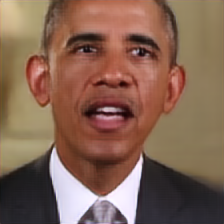}
        \\
        \centering\rotatebox{90}{\textbf{Ours}}&
        \includegraphics[align=c,bmargin=0.13cm,width=\wid,trim={20px 22px 20px 18px},clip]{figures_rebuttal/f2f/f2f_ours_7.png}
        \;&\;
        \includegraphics[align=c,width=\wid,trim={20px 22px 20px 18px},clip]{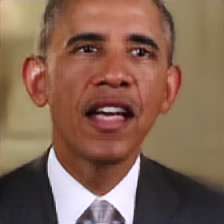}&
        \includegraphics[align=c,width=\wid,trim={20px 22px 20px 18px},clip]{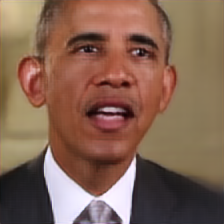}&
        \includegraphics[align=c,width=\wid,trim={20px 22px 20px 18px},clip]{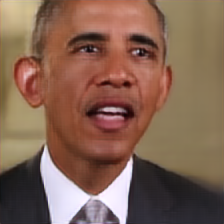}&
        \includegraphics[align=c,width=\wid,trim={20px 22px 20px 18px},clip]{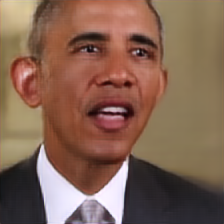}&
        \includegraphics[align=c,width=\wid,trim={20px 22px 20px 18px},clip]{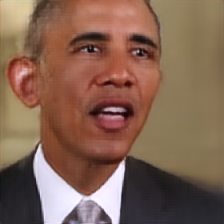}
        \\
        & & \multicolumn{5}{c}{\textbf{Ours, multi-view synthesis}}
    \end{tabular}
    \caption{Comparison with Thies et al.\cite{Thies16}. We used 32 frames for the fine-tuning, while 1100 frames were used to train the Face2Face model. Note that the output resolution of our model is constrained by the training dataset. Also, our model is able to synthesize a naturally looking frame from different viewpoints for a fixed pose (given 3D face landmarks), which is a limitation of the Face2Face system.}
    \label{fig:f2f}
\end{figure*}

\subsection{Ablation study}

In this section, we evaluate the contributions related to the losses we use in the training of our model, as well as motivate the training procedure. We have already shown in \fig{voxceleb2} the effect that the fine-tuning has on the quality of the results, so we do not evaluate it here. Instead, we focus on the details of fine-tuning. 

The first question we asked was about the importance of person-specific parameters initialization via the embedder. We tried different types of random initialization for both the embedding vector $\hat\e\new$ and the adaptive parameters $\hat\psi$ of the generator, but these experiments did not yield any plausible images after the fine-tuning. Hence we realized that the person-specific initialization of the generator provided by the embedder is important for convergence of the fine-tuning problem.

\begin{figure*}
    \centering
    \setlength{\wid}{0.179\textwidth}
    \addtolength{\tabcolsep}{-4pt}
    \begin{tabular}{m{0.6cm}c:ccccc}
        \centering{\textbf{1}}&
        \includegraphics[align=c,bmargin=0.13cm,width=\wid]{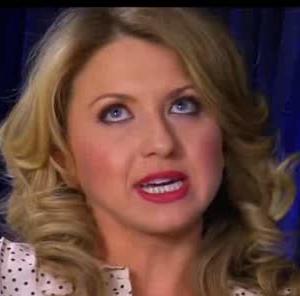}
        \;&\;
        \includegraphics[align=c,width=\wid]{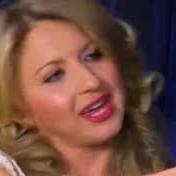}&
        \includegraphics[align=c,width=\wid]{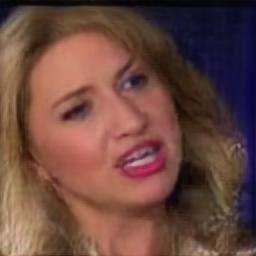}&
        \includegraphics[align=c,width=\wid]{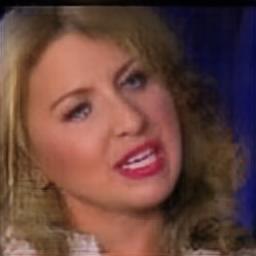}&
        \includegraphics[align=c,width=\wid]{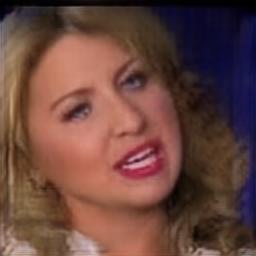}\\
        \centering{\textbf{8}}&
        \includegraphics[align=c,bmargin=0.13cm,width=\wid]{figures_supmat/figure6/0_src.jpg}
        \;&\;
        \includegraphics[align=c,width=\wid]{figures_supmat/figure6/0_gt.jpg}&
        \includegraphics[align=c,width=\wid]{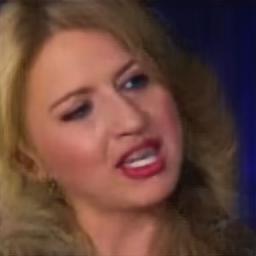}&
        \includegraphics[align=c,width=\wid]{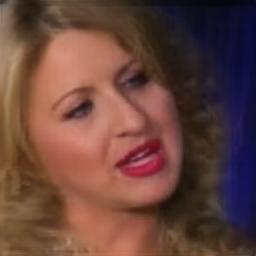}&
        \includegraphics[align=c,width=\wid]{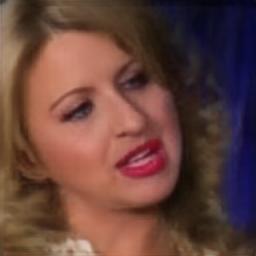}\\
        \centering{\textbf{32}}&
        \includegraphics[align=c,bmargin=0.13cm,width=\wid]{figures_supmat/figure6/0_src.jpg}
        \;&\;
        \includegraphics[align=c,width=\wid]{figures_supmat/figure6/0_gt.jpg}&
        \includegraphics[align=c,width=\wid]{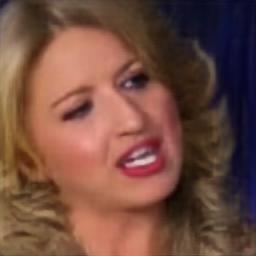}&
        \includegraphics[align=c,width=\wid]{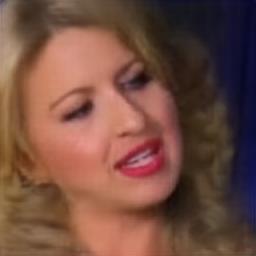}&
        \includegraphics[align=c,width=\wid]{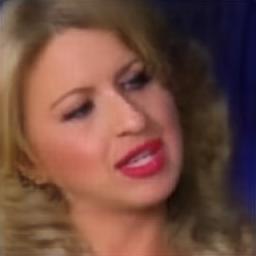}\\
        \centering{\textbf{1}}&
        \includegraphics[align=c,bmargin=0.13cm,width=\wid]{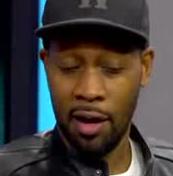}
        \;&\;
        \includegraphics[align=c,width=\wid]{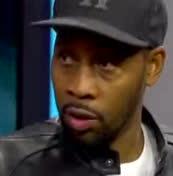}&
        \includegraphics[align=c,width=\wid]{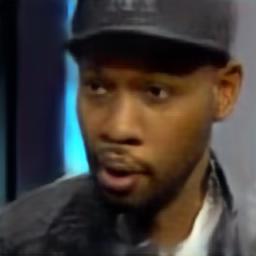}&
        \includegraphics[align=c,width=\wid]{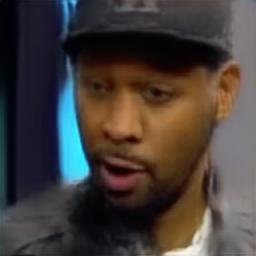}&
        \includegraphics[align=c,width=\wid]{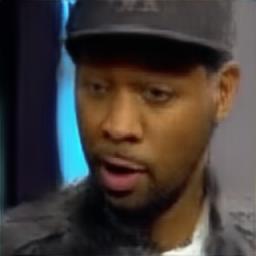}\\
        \centering{\textbf{8}}&
        \includegraphics[align=c,bmargin=0.13cm,width=\wid]{figures_supmat/figure6/1_src.jpg}
        \;&\;
        \includegraphics[align=c,width=\wid]{figures_supmat/figure6/1_gt.jpg}&
        \includegraphics[align=c,width=\wid]{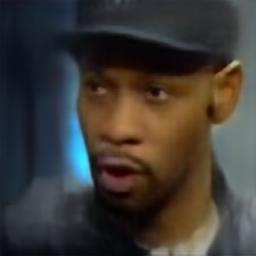}&
        \includegraphics[align=c,width=\wid]{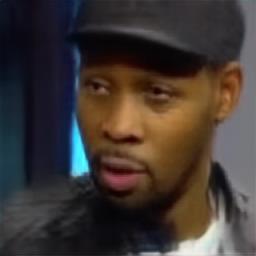}&
        \includegraphics[align=c,width=\wid]{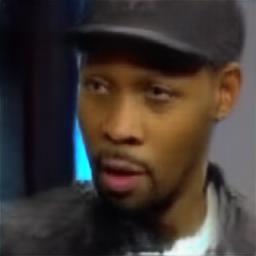}\\
        \centering{\textbf{32}}&
        \includegraphics[align=c,bmargin=0.13cm,width=\wid]{figures_supmat/figure6/1_src.jpg}
        \;&\;
        \includegraphics[align=c,width=\wid]{figures_supmat/figure6/1_gt.jpg}&
        \includegraphics[align=c,width=\wid]{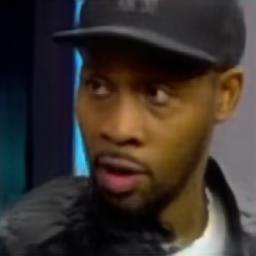}&
        \includegraphics[align=c,width=\wid]{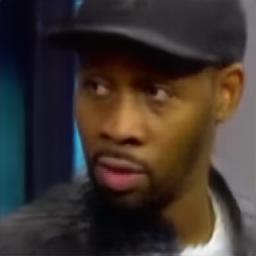}&
        \includegraphics[align=c,width=\wid]{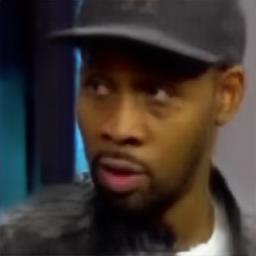}\\
        & \textbf{Source} \;&\; \textbf{Ground truth} & \textbf{\begin{tabular}{c}w/o $\mtch$,\\ random $\W_i$\end{tabular}} & \textbf{w/o $\aadv$} & \textbf{Ours}
    \end{tabular}
    \captionof{figure}{Ablation study of our contributions. The number of training frames is, again, equal to \textbf{T} (the leftmost column), the example training frame in shown in the \textbf{source} column and the next column shows \textbf{ground truth} image. Then, we remove \textbf{$\mtch$} from the meta-learning objective and initialize the embedding vector of the discriminator \textbf{randomly} (third column) and evaluate the contribution of adversarial fine-tuning compared to the regular fine-tuning \textbf{with no $\aadv$} in the objective (fifth column). The last column represents results from our final model.}
    \label{fig:ablationsuppmat}
\end{figure*}

Then, we evaluated the contribution of the person-specific initialization of the discriminator. We remove $\mtch$ term from the objective and perform meta-learning. The use of multiple training frames in few-shot learning problems, like in our final method, leads to optimization instabilities, so we used a one-shot meta-learning configuration, which turned out to be stable. After meta-learning, we randomly initialize the person-specific vector $\W_i$ of the discriminator. The results can be seen in \fig{ablationsuppmat}. We notice that the results for random initialization are plausible but introduce a noticeable gap in terms of realism and personalization fidelity. We, therefore, came to the conclusion that person-specific initialization of the discriminator also contributes to the quality of the results, albeit in a lesser way than the initialization of the generator does.

\begin{figure*}
    \centering
    \setlength{\wid}{0.14\textwidth}
    \addtolength{\tabcolsep}{-4pt}
    \begin{tabular}{m{0.6cm}c:cc c:cc}
        &
        \includegraphics[align=c,bmargin=0.13cm,width=\wid,trim={30px 136px 106px 0px},clip]{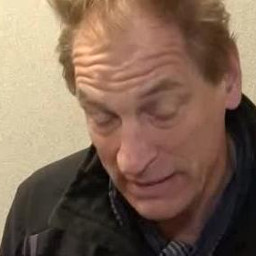}
        \;&\;
        \includegraphics[align=c,width=\wid,trim={30px 136px 106px 0px},clip]{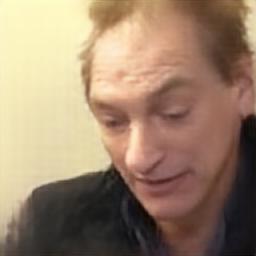}&
        \includegraphics[align=c,width=\wid,trim={30px 136px 106px 0px},clip]{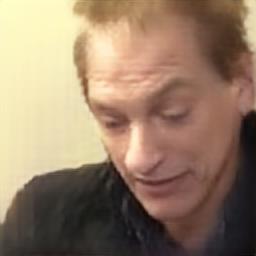}
        \;&\;
        \includegraphics[align=c,width=\wid,trim={0px 136px 136px 0px},clip]{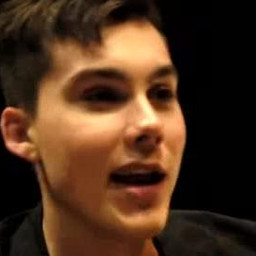}
        \;&\;
        \includegraphics[align=c,width=\wid,trim={0px 136px 136px 0px},clip]{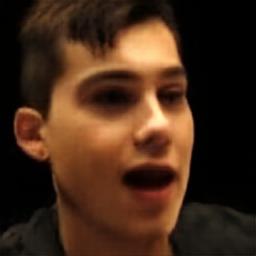}&
        \includegraphics[align=c,width=\wid,trim={0px 136px 136px 0px},clip]{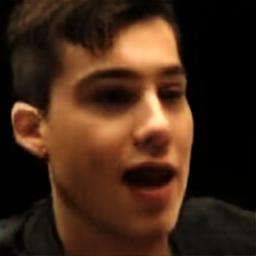}
        \\
        &
        \includegraphics[align=c,bmargin=0.13cm,width=\wid,trim={40px 106px 116px 50px},clip]{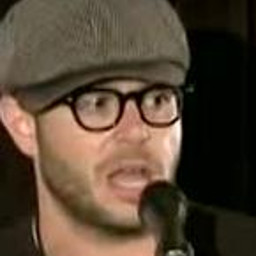}
        \;&\;
        \includegraphics[align=c,width=\wid,trim={40px 106px 116px 50px},clip]{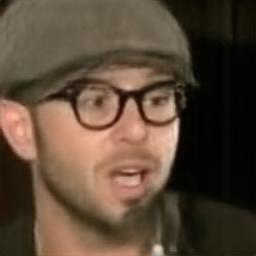}&
        \includegraphics[align=c,width=\wid,trim={40px 106px 116px 50px},clip]{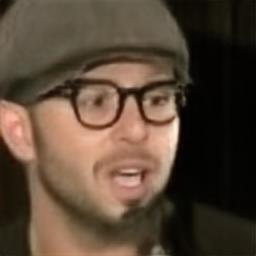}
        \;&\;
        \includegraphics[align=c,width=\wid,trim={171px 30px 0px 141px},clip]{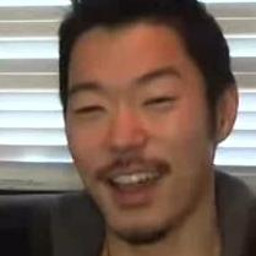}
        \;&\;
        \includegraphics[align=c,width=\wid,trim={171px 30px 0px 141px},clip]{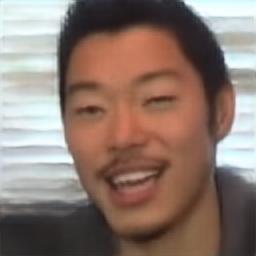}&
        \includegraphics[align=c,width=\wid,trim={171px 30px 0px 141px},clip]{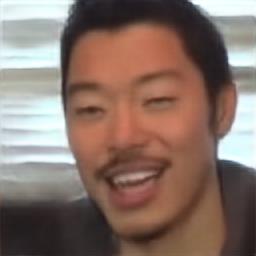}
        \\
        & \textbf{Source} & \textbf{w/0 $\aadv$} & \textbf{Ours} & \textbf{Source} & \textbf{w/o $\aadv$} & \textbf{Ours}
    \end{tabular}
    \caption{More close-up examples of the ablation study examples for the comparison against the model \textbf{w/o $\aadv$}. We used 8 training frames. Notice the geometry gap (top row) and additional artifacts (bottom row) introduced by the removal of $\aadv$ during fine-tuning.}
    \label{fig:ablationcloseups}
\end{figure*}

Finally, we evaluate the contribution of adversarial term $\aadv$ during the fine-tuning. We, therefore, remove it from the fine-tuning objective and compare the results to our best model (see \fig{ablationsuppmat}). While the difference between these variants is quite subtle, we note that adversarial fine-tuning leads to crisper images that better match ground truth both in terms of pose and image details. The close-up images in \fig{ablationcloseups} were chosen in order to highlight these differences.

\subsection{Additional qualitative results}

\begin{figure*}
    \centering
    \setlength{\wid}{0.14\textwidth}
    \addtolength{\tabcolsep}{-4pt}
    \begin{tabular}{m{0.6cm} c:cc c:cc}
        \centering\rotatebox{90}{\textbf{Driver}}&
        \includegraphics[align=c,bmargin=0.13cm,width=\wid,trim={0 0 0 112px},clip]{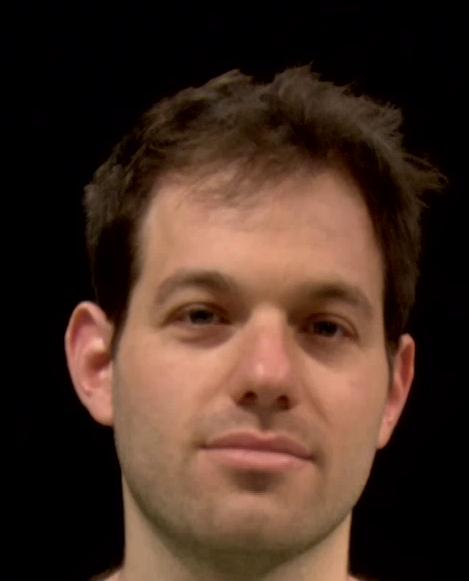}
        \;&\;
        \includegraphics[align=c,width=\wid,trim={0 0 0 112px},clip]{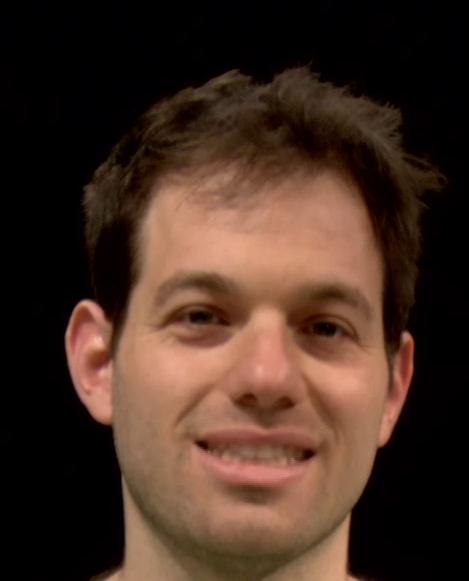}&
        \includegraphics[align=c,width=\wid,trim={0 0 0 112px},clip]{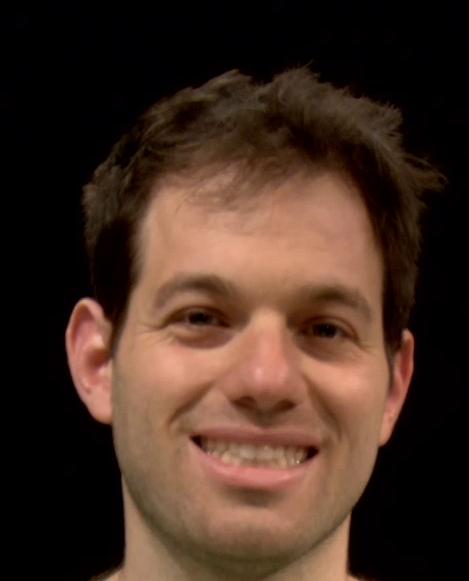}
        \;&\;
        \includegraphics[align=c,width=\wid,trim={0 0 0 112px},clip]{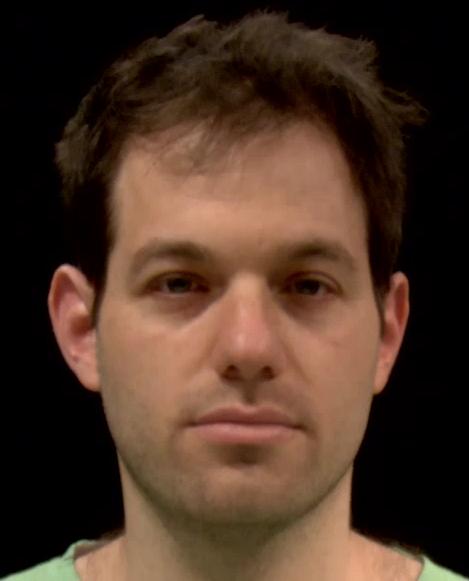}
        \;&\;
        \includegraphics[align=c,width=\wid,trim={0 0 0 112px},clip]{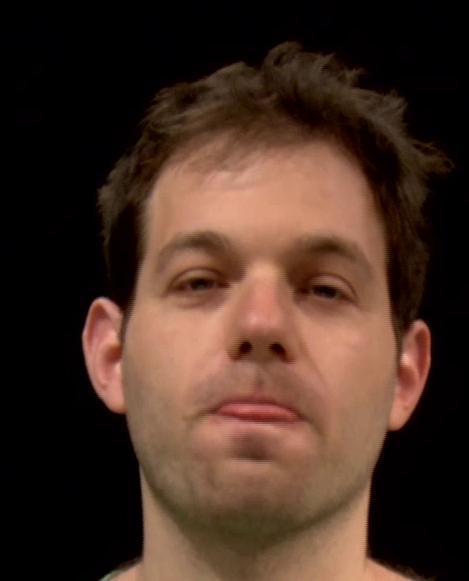}&
        \includegraphics[align=c,width=\wid,trim={0 0 0 112px},clip]{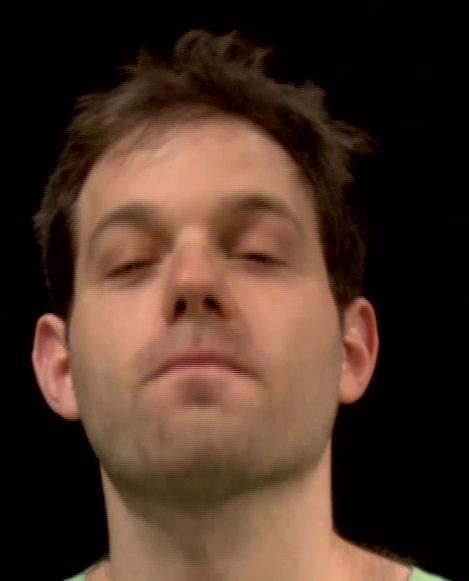}
        \\
        \centering\rotatebox{90}{\textbf{Averbuch et al.}}&
        \includegraphics[align=c,bmargin=0.13cm,width=\wid,trim={0 0 0 112px},clip]{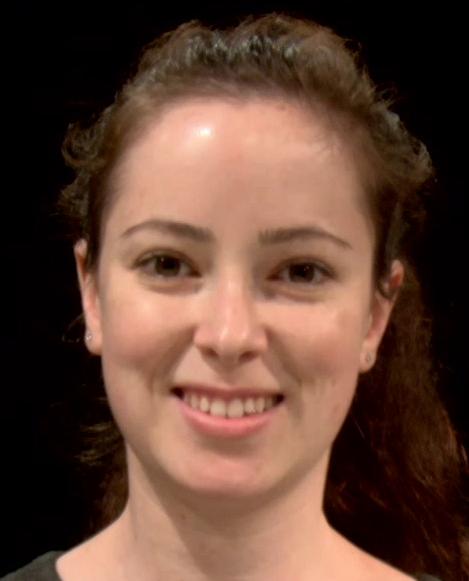}
        \;&\;
        \includegraphics[align=c,width=\wid,trim={0 0 0 112px},clip]{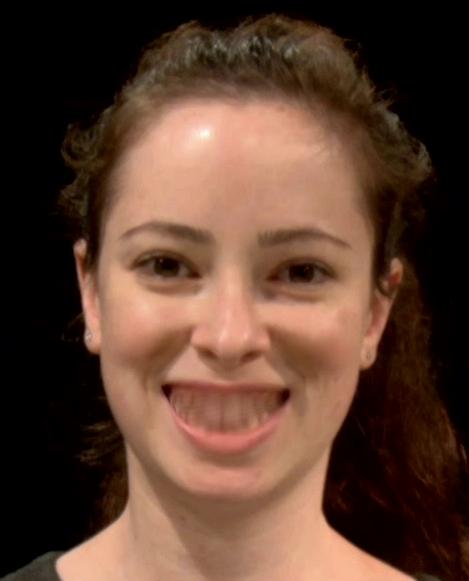}&
        \includegraphics[align=c,width=\wid,trim={0 0 0 112px},clip]{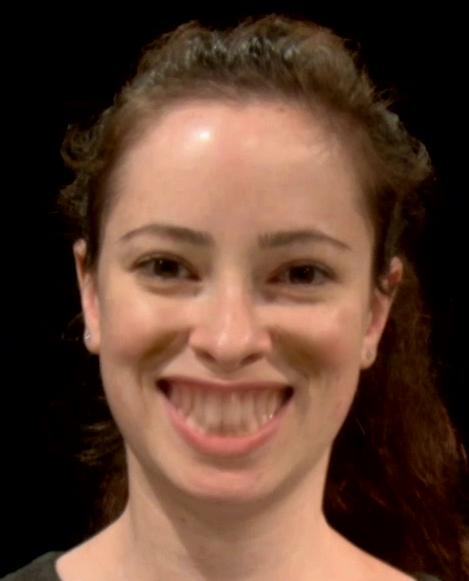}
        \;&\;
        \includegraphics[align=c,width=\wid,trim={0 0 0 112px},clip]{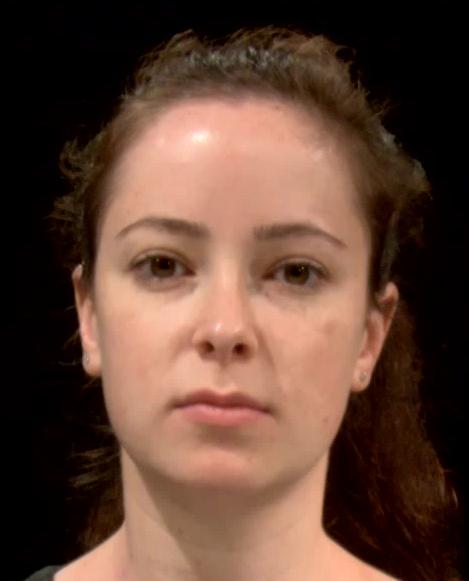}
        \;&\;
        \includegraphics[align=c,width=\wid,trim={0 0 0 112px},clip]{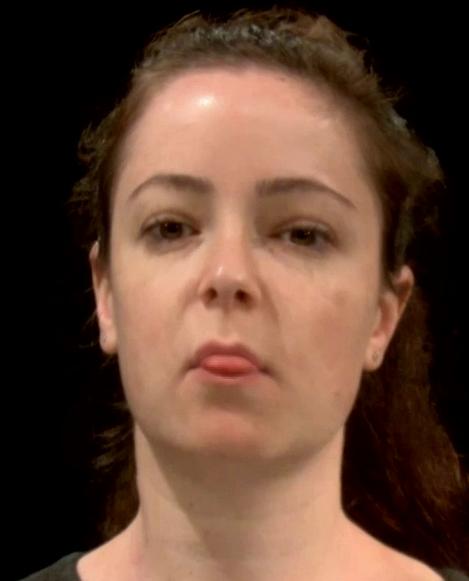}&
        \includegraphics[align=c,width=\wid,trim={0 0 0 112px},clip]{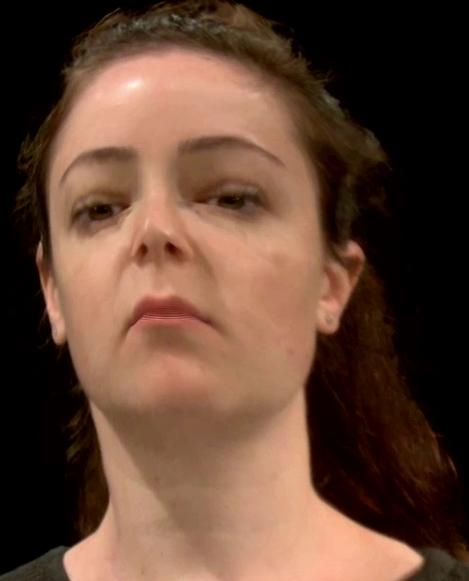}
        \\
        \centering\rotatebox{90}{\textbf{Ours}}&
        \includegraphics[align=c,bmargin=0.13cm,width=\wid]{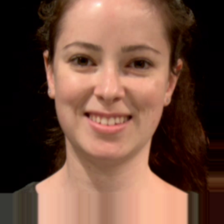}
        \;&\;
        \includegraphics[align=c,width=\wid]{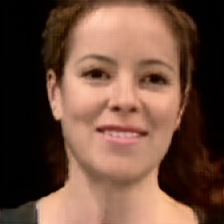}&
        \includegraphics[align=c,width=\wid]{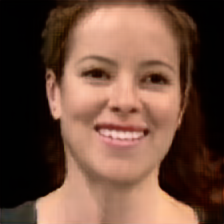}
        \;&\;
        \includegraphics[align=c,width=\wid]{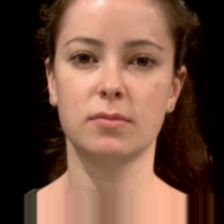}
        \;&\;
        \includegraphics[align=c,width=\wid]{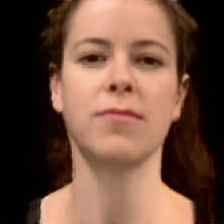}&
        \includegraphics[align=c,width=\wid]{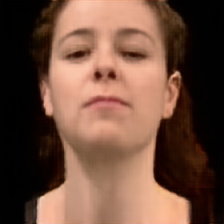}
        \\
        & \textbf{Source} & \multicolumn{2}{c}{\textbf{Results}} & \textbf{Source} & \multicolumn{2}{c}{\textbf{Results}}
    \end{tabular}
    \caption{Comparison with Averbuch-Elor et al.~\cite{Averbuch17} on the failure cases mentioned in the paper. Notice that our model better transfers the input pose and also is unaffected by the pose of the original frame, which lifts the "neutral face" constraint on the source image assumed in~\cite{Averbuch17}.}
    \label{fig:bpl}
\end{figure*}

\begin{figure*}
    \centering
    \setlength{\wid}{0.143\textwidth}
    \addtolength{\tabcolsep}{-4pt}
    \begin{tabular}{m{0.6cm} c : c : cccc}
        &
        \includegraphics[align=c,bmargin=0.13cm,width=\wid,trim={0px 0px 7px 0px},clip]{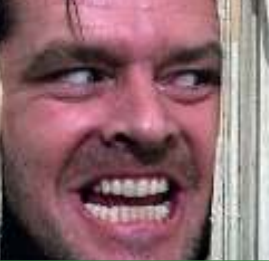}
        \;&\;
        \includegraphics[align=c,width=\wid]{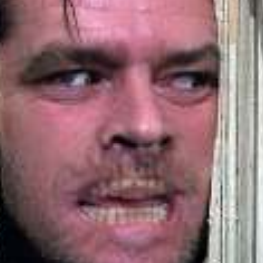}
        \;&\;
        \includegraphics[align=c,width=\wid,trim={65px 90px 45px 20px},clip]{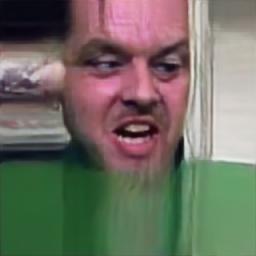}&
        \includegraphics[align=c,width=\wid,trim={65px 90px 45px 20px},clip]{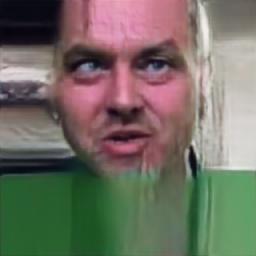}&
        \includegraphics[align=c,width=\wid,trim={65px 90px 45px 20px},clip]{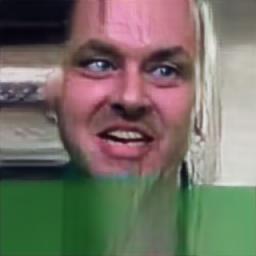}&
        \includegraphics[align=c,width=\wid,trim={65px 90px 45px 20px},clip]{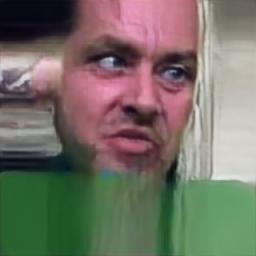}
        \\
        & \textbf{Source} & \textbf{GANimation} & \multicolumn{4}{c}{\textbf{Our driving results}}
    \end{tabular}\vspace{-9pt}
    \caption{Comparison with Pumarola et al.~\cite{Pumarola18} (second column) and our method (right four columns). We perform the driving in the same way as we animate still images in the paper. Note that in the VoxCeleb datasets face cropping have been performed differently, so we had to manually crop our results, effectively decreasing the resolution.}
    \label{fig:ganimation}
\end{figure*}

\begin{figure*}
    \centering
    \setlength{\wid}{0.179\textwidth}
    \addtolength{\tabcolsep}{-4pt}
    \begin{tabular}{m{0.6cm}c:cccc}
        &
        \includegraphics[width=\wid]{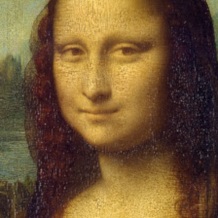}
        \;&\;
        \includegraphics[width=\wid]{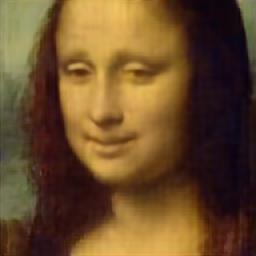}&
        \includegraphics[width=\wid]{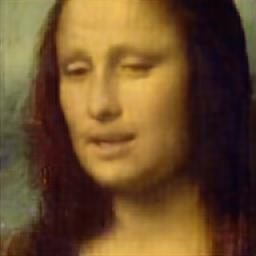}&
        \includegraphics[width=\wid]{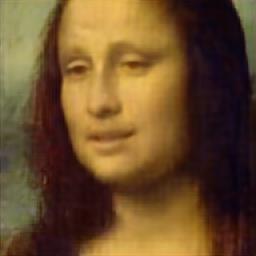}&
        \includegraphics[width=\wid]{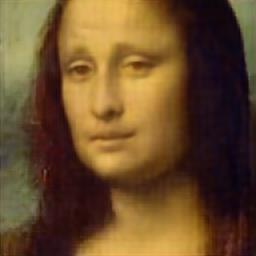}\\
        &
        \includegraphics[width=\wid]{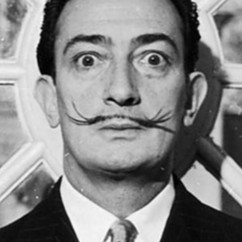}
        \;&\;
        \includegraphics[width=\wid]{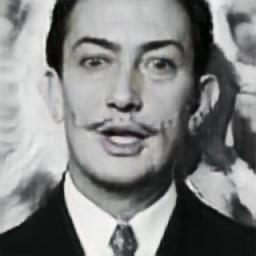}&
        \includegraphics[width=\wid]{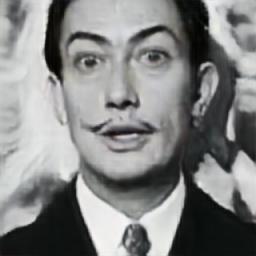}&
        \includegraphics[width=\wid]{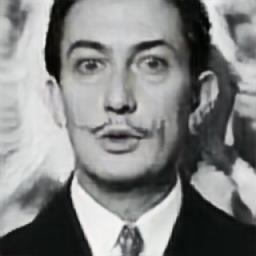}&
        \includegraphics[width=\wid]{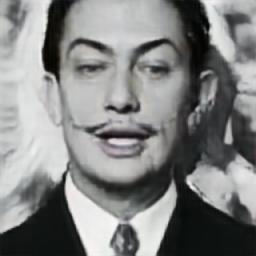}\\
        &
        \includegraphics[width=\wid]{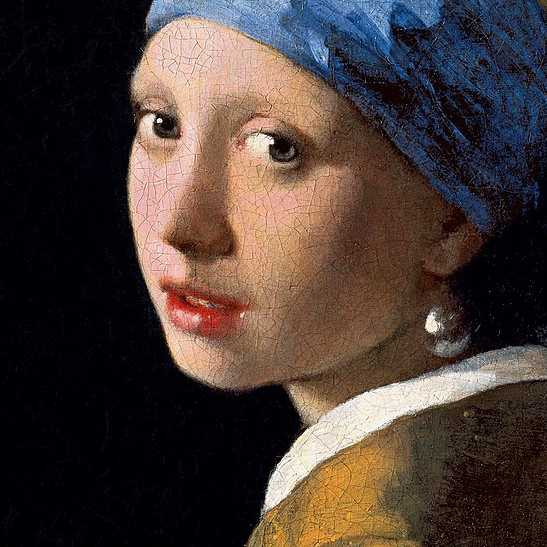}
        \;&\;
        \includegraphics[width=\wid]{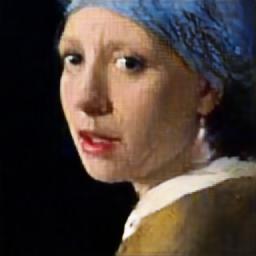}&
        \includegraphics[width=\wid]{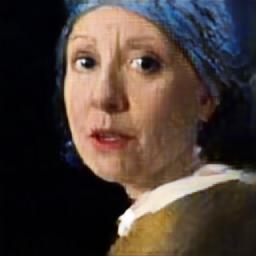}&
        \includegraphics[width=\wid]{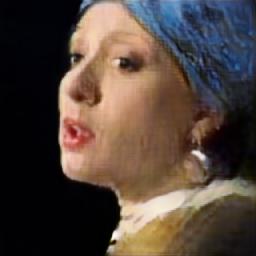}&
        \includegraphics[width=\wid]{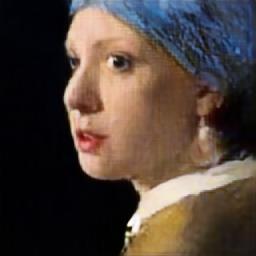}\\
        &
        \includegraphics[width=\wid]{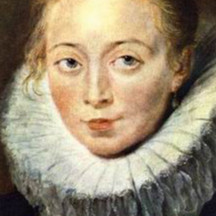}
        \;&\;
        \includegraphics[width=\wid]{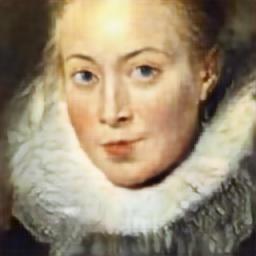}&
        \includegraphics[width=\wid]{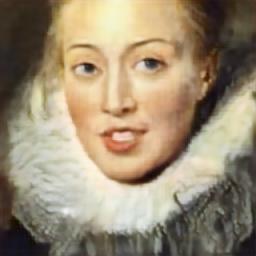}&
        \includegraphics[width=\wid]{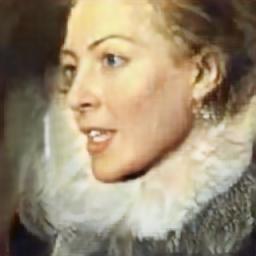}&
        \includegraphics[width=\wid]{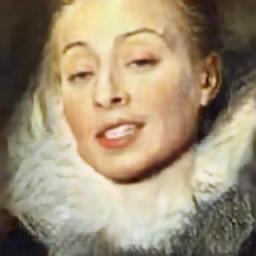}\\
        &
        \includegraphics[width=\wid]{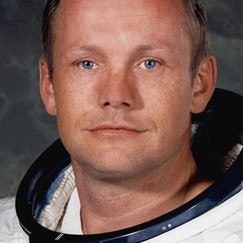}
        \;&\;
        \includegraphics[width=\wid]{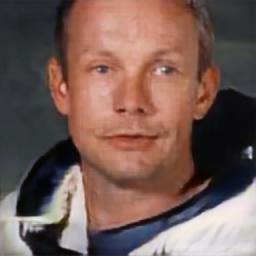}&
        \includegraphics[width=\wid]{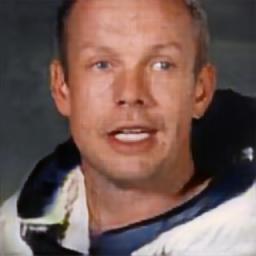}&
        \includegraphics[width=\wid]{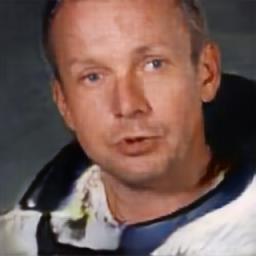}&
        \includegraphics[width=\wid]{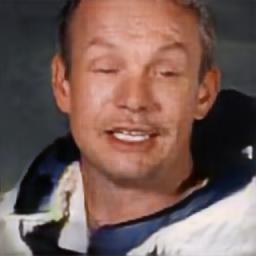}\\
        &
        \includegraphics[width=\wid]{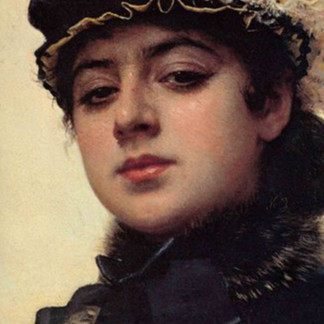}
        \;&\;
        \includegraphics[width=\wid]{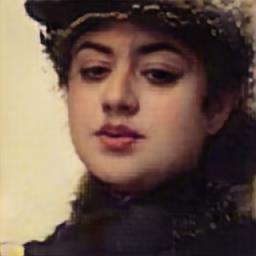}&
        \includegraphics[width=\wid]{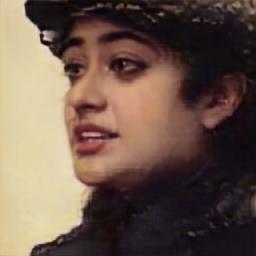}&
        \includegraphics[width=\wid]{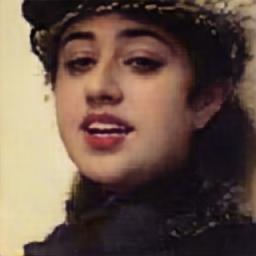}&
        \includegraphics[width=\wid]{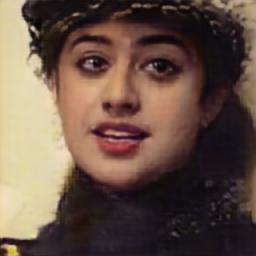}\\ 
        & \textbf{Source} & \multicolumn{4}{c}{\textbf{Generated images}}
    \end{tabular}
    \captionof{figure}{More puppeteering results for talking head models trained in one-shot setting. The image used for one-shot training problem is in the \textbf{source} column. The next columns show \textbf{generated images}, which were conditioned on the video sequence of a different person.}
    \label{fig:livingportraitssuppmat}
\end{figure*}

\begin{figure*}
    \centering
    \setlength{\wid}{0.179\textwidth}
    \addtolength{\tabcolsep}{-4pt}
    \begin{tabular}{m{0.6cm}c:cccc}
        &
        \includegraphics[width=\wid]{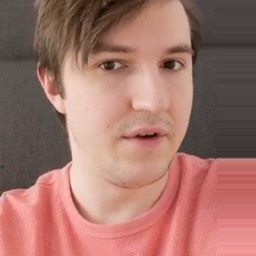}
        \;&\;
        \includegraphics[width=\wid]{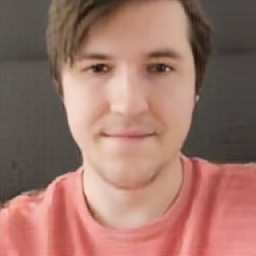}&
        \includegraphics[width=\wid]{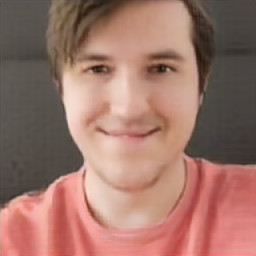}&
        \includegraphics[width=\wid]{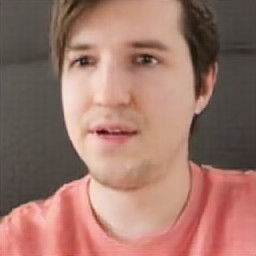}&
        \includegraphics[width=\wid]{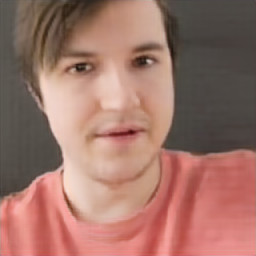}\\
        &
        \includegraphics[width=\wid]{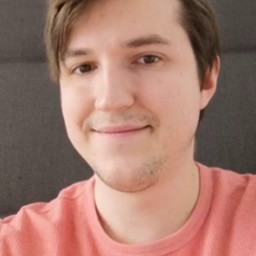}
        \;&\;
        \includegraphics[width=\wid]{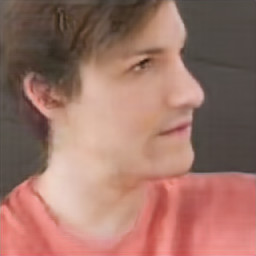}&
        \includegraphics[width=\wid]{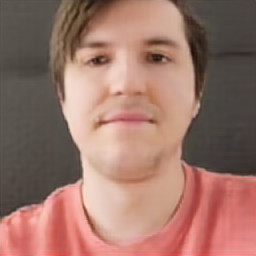}&
        \includegraphics[width=\wid]{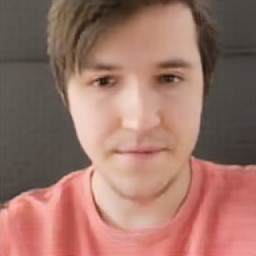}&
        \includegraphics[width=\wid]{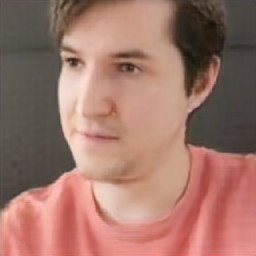}\\
        &
        \includegraphics[width=\wid]{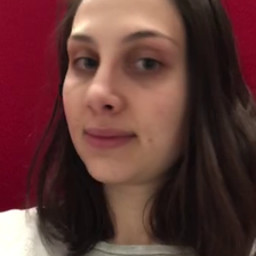}
        \;&\;
        \includegraphics[width=\wid]{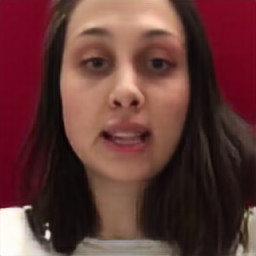}&
        \includegraphics[width=\wid]{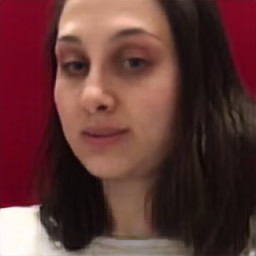}&
        \includegraphics[width=\wid]{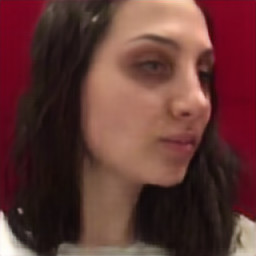}&
        \includegraphics[width=\wid]{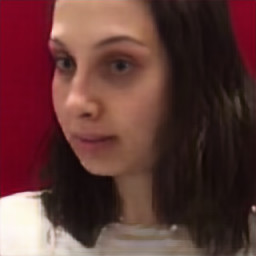}\\
        &
        \includegraphics[width=\wid]{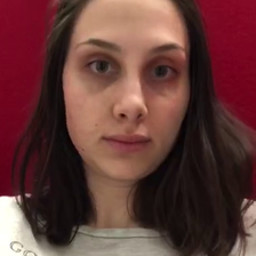}
        \;&\;
        \includegraphics[width=\wid]{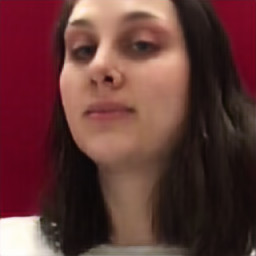}&
        \includegraphics[width=\wid]{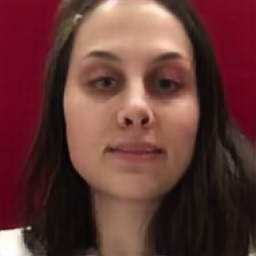}&
        \includegraphics[width=\wid]{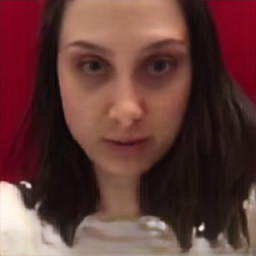}&
        \includegraphics[width=\wid]{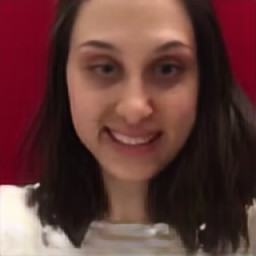}\\
        &
        \includegraphics[width=\wid]{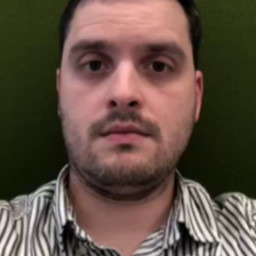}
        \;&\;
        \includegraphics[width=\wid]{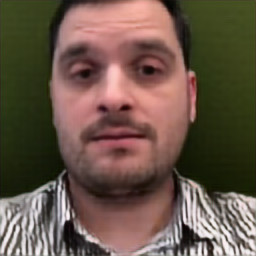}&
        \includegraphics[width=\wid]{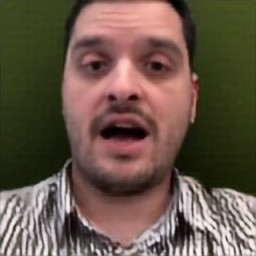}&
        \includegraphics[width=\wid]{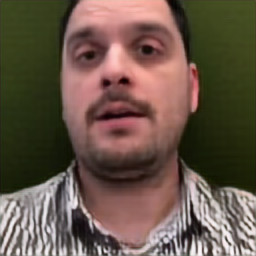}&
        \includegraphics[width=\wid]{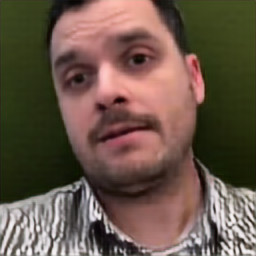}\\
        &
        \includegraphics[width=\wid]{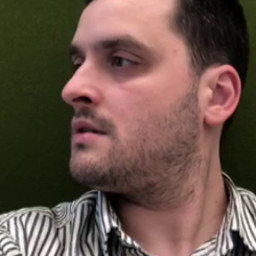}
        \;&\;
        \includegraphics[width=\wid]{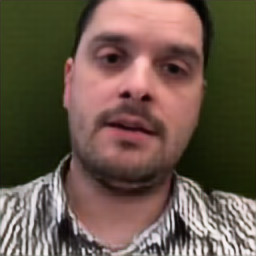}&
        \includegraphics[width=\wid]{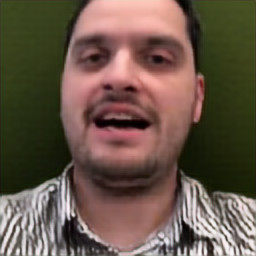}&
        \includegraphics[width=\wid]{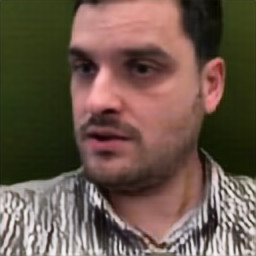}&
        \includegraphics[width=\wid]{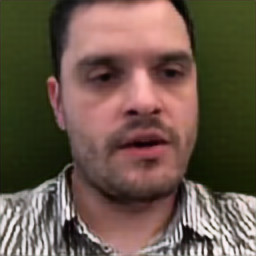}\\ 
        & \textbf{Source} & \multicolumn{4}{c}{\textbf{Generated images}}
    \end{tabular}
    \captionof{figure}{Results for talking head models trained in 16-shot setting on selfie photographs with driving landmarks taken from the different video of the same person. Example training frames are shown in the \textbf{source} column. The next columns show \textbf{generated images}, which were conditioned on the different video sequence of the same person.}
    \label{fig:selfiessuppmat}
\end{figure*}

More comparisons with other methods are available in \fig{bpl}, \fig{ganimation}, \fig{f2f}. More puppeteering results for one-shot learned portraits and photographs are presented in \fig{livingportraitssuppmat}. We also show the results for talking heads learned from selfies in \fig{selfiessuppmat}. Additional comparisons between the methods are provided in the rest of the figures.

\subsection{Training and architecture details}

As stated in the paper, we used the architecture similar to the one in~\cite{Brock18}. The convolutional parts of the embedder and the discriminator are the same networks with 6 residual downsampling blocks, each performing downsampling by a factor of 2. The inputs of these convolutional networks are RGB images concatenated with the landmark images, in total there are 6 input channels. The initial number of channels is 64, increased by a factor of two in each block, up to a maximum of 512. The blocks are pre-activated residual blocks with no normalization, as described in the paper~\cite{Brock18}. The first block is a regular residual block with activation function not being applied in the end. Each skip connection has a linear layer inside if the spatial resolution is being changed. Self-attention~\cite{Zhang18b} blocks are inserted after three downsampling blocks. Downsampling is performed via average pooling. Then, after applying ReLU activation function to the output tensor, we perform sum-pooling over spatial dimensions.

For the embedder, the resulting vectorized embeddings for each training image are stored (in order to apply $\mtch$ element-wise), and the averaged embeddings are fed into the generator. For the discriminator, the resulting vector is used to calculate the realism score.

The generator consists of three parts: 4 residual downsampling blocks (with self-attention inserted before the last block), 4 blocks operating at bottleneck resolution and 4 upsampling blocks (self-attention is inserted after 2 upsampling blocks). Upsampling is performed in the end of the block, following~\cite{Brock18}. The number of channels in bottleneck layers is 512. Downsampling blocks are normalized via instance normalization~\cite{Ulyanov16}, while bottleneck and upsampling blocks are normalized via adaptive instance normalization. A single linear layer is used to map an embedding vector to all adaptive parameters. After the last upsampling block, we insert a final adaptive normalization layer, followed by a ReLU and a convolution. The output is then mapped into $[-1, 1]$ via Tanh.

The training was carried out on 8 NVIDIA P40 GPUs, with batch size 48 via simultaneous gradient descend, with 2 updates of the discriminator per 1 of the generator. In our experiments, we used PyTorch distributed module and have performed reduction of the gradients across the GPUs only for the generator and the embedder.

\begin{figure*}
    \centering
    \setlength{\wid}{0.179\textwidth}
    \addtolength{\tabcolsep}{-4pt}
    \begin{tabular}{m{0.6cm}c:cccc}
        &
        \includegraphics[width=\wid]{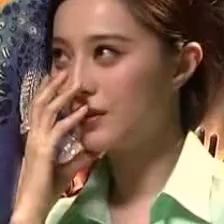}
        \;&\;
        \includegraphics[width=\wid]{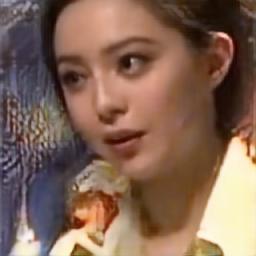}&
        \includegraphics[width=\wid]{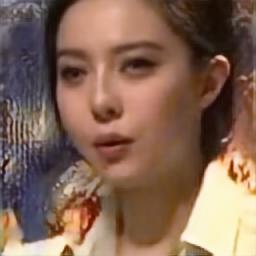}&
        \includegraphics[width=\wid]{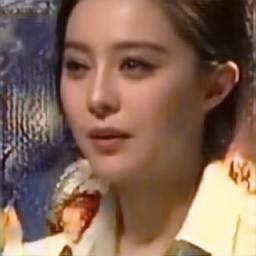}&
        \includegraphics[width=\wid]{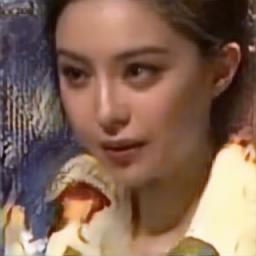}\\
        &
        \includegraphics[width=\wid]{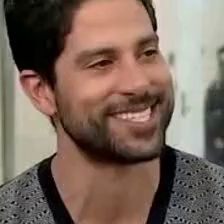}
        \;&\;
        \includegraphics[width=\wid]{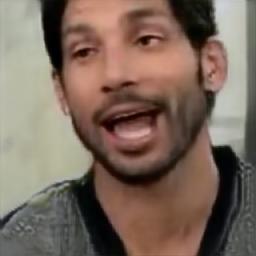}&
        \includegraphics[width=\wid]{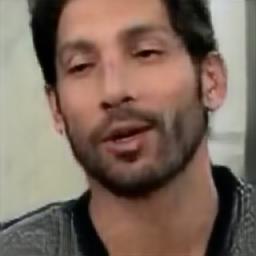}&
        \includegraphics[width=\wid]{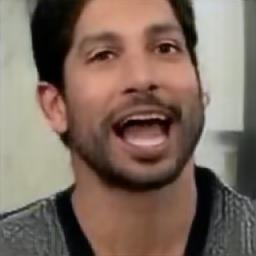}&
        \includegraphics[width=\wid]{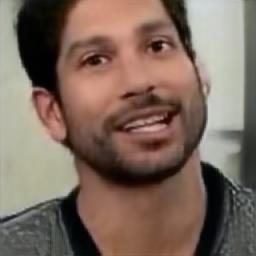}\\
        &
        \includegraphics[width=\wid]{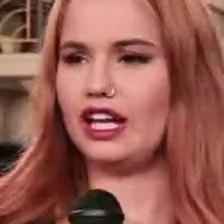}
        \;&\;
        \includegraphics[width=\wid]{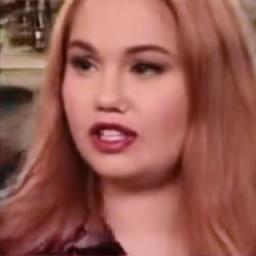}&
        \includegraphics[width=\wid]{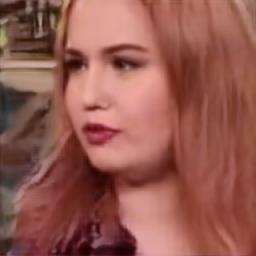}&
        \includegraphics[width=\wid]{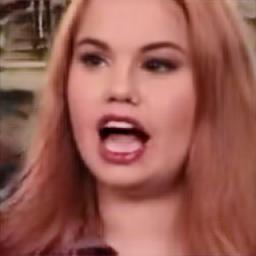}&
        \includegraphics[width=\wid]{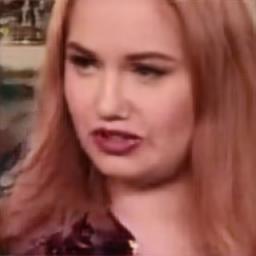}\\
        &
        \includegraphics[width=\wid]{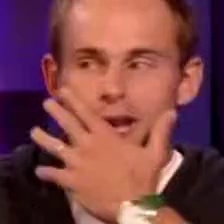}
        \;&\;
        \includegraphics[width=\wid]{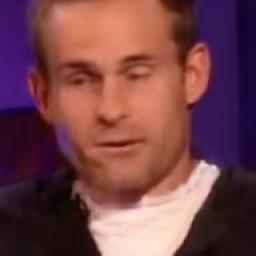}&
        \includegraphics[width=\wid]{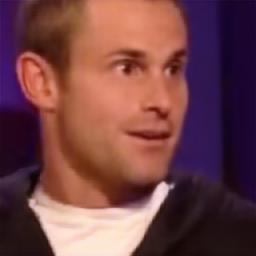}&
        \includegraphics[width=\wid]{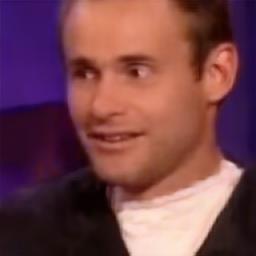}&
        \includegraphics[width=\wid]{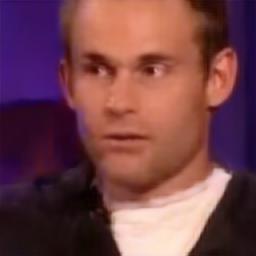}\\
        &
        \includegraphics[width=\wid]{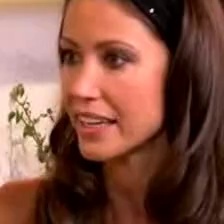}
        \;&\;
        \includegraphics[width=\wid]{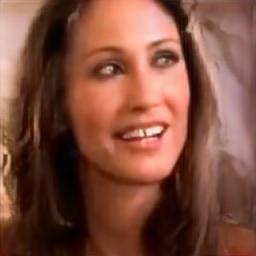}&
        \includegraphics[width=\wid]{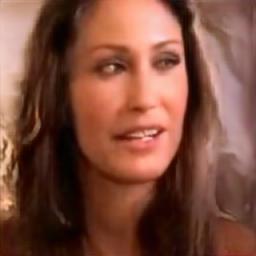}&
        \includegraphics[width=\wid]{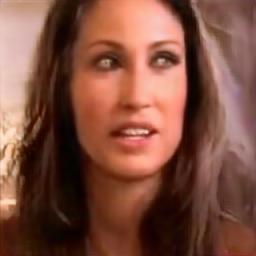}&
        \includegraphics[width=\wid]{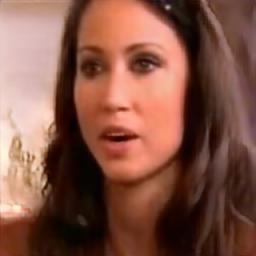}\\
        &
        \includegraphics[width=\wid]{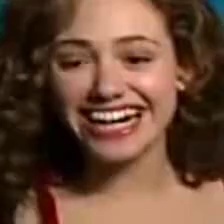}
        \;&\;
        \includegraphics[width=\wid]{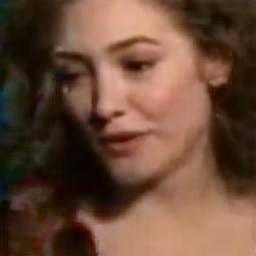}&
        \includegraphics[width=\wid]{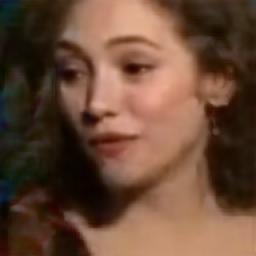}&
        \includegraphics[width=\wid]{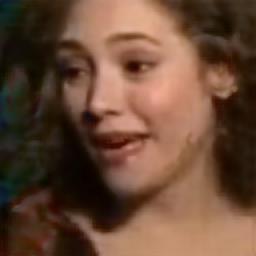}&
        \includegraphics[width=\wid]{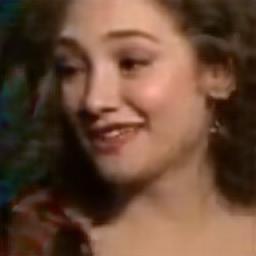}\\ 
        & \textbf{Source} & \multicolumn{4}{c}{\textbf{Generated images}}
    \end{tabular}
    \captionof{figure}{Results for talking head models trained in eight-shot setting. Example training frame is in the \textbf{source} column. The next columns show \textbf{generated images}, which were conditioned on the pose tracks taken from a different video sequence with the same person.}
    \label{fig:driversuppmat}
\end{figure*}

\begin{figure*}
    \centering    
    \setlength{\wid}{0.179\textwidth}
    \addtolength{\tabcolsep}{-4pt}
    \begin{tabular}{m{0.6cm}c:cccc}
        \centering{\textbf{1}}&
        \includegraphics[align=c,bmargin=0.13cm,width=\wid]{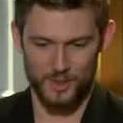}
        \;&\;
        \includegraphics[align=c,width=\wid]{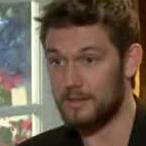}&
        \includegraphics[align=c,width=\wid]{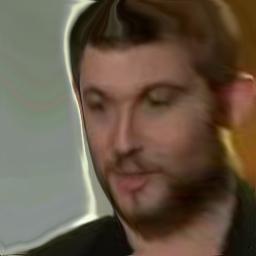}&
        \includegraphics[align=c,width=\wid]{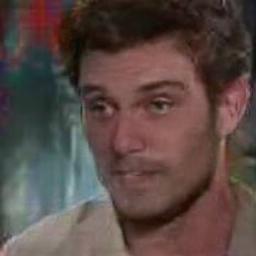}&
        \includegraphics[align=c,width=\wid]{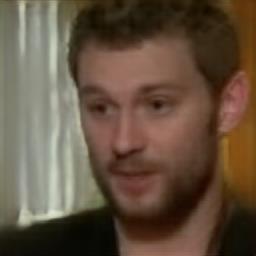}\\
        \centering{\textbf{8}}&
        \includegraphics[align=c,bmargin=0.13cm,width=\wid]{figures_supmat/figure10/0_src.jpg}
        \;&\;
        \includegraphics[align=c,width=\wid]{figures_supmat/figure10/0_gt.jpg}&
        \includegraphics[align=c,width=\wid]{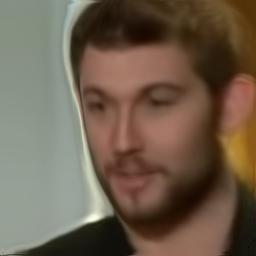}&
        \includegraphics[align=c,width=\wid]{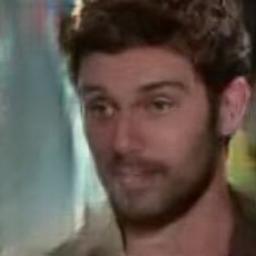}&
        \includegraphics[align=c,width=\wid]{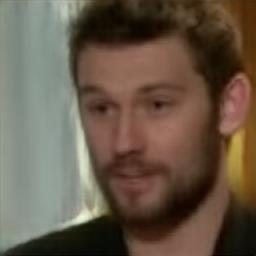}\\
        \centering{\textbf{32}}&
        \includegraphics[align=c,bmargin=0.13cm,width=\wid]{figures_supmat/figure10/0_src.jpg}
        \;&\;
        \includegraphics[align=c,width=\wid]{figures_supmat/figure10/0_gt.jpg}&
        \includegraphics[align=c,width=\wid]{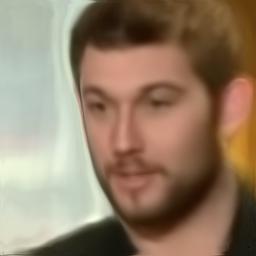}&
        \includegraphics[align=c,width=\wid]{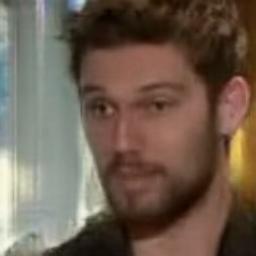}&
        \includegraphics[align=c,width=\wid]{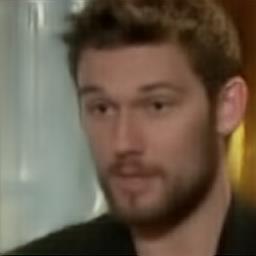}\\
        \centering{\textbf{1}}&
        \includegraphics[align=c,bmargin=0.13cm,width=\wid]{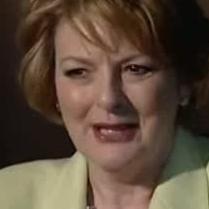}
        \;&\;
        \includegraphics[align=c,width=\wid]{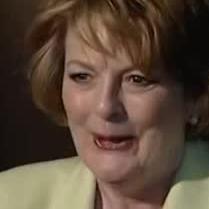}&
        \includegraphics[align=c,width=\wid]{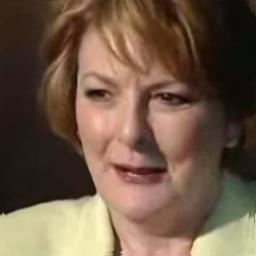}&
        \includegraphics[align=c,width=\wid]{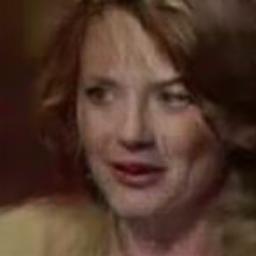}&
        \includegraphics[align=c,width=\wid]{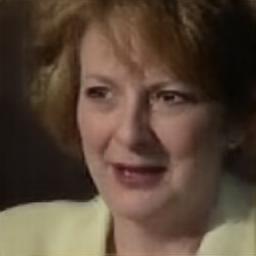}\\
        \centering{\textbf{8}}&
        \includegraphics[align=c,bmargin=0.13cm,width=\wid]{figures_supmat/figure10/1_src.jpg}
        \;&\;
        \includegraphics[align=c,width=\wid]{figures_supmat/figure10/1_gt.jpg}&
        \includegraphics[align=c,width=\wid]{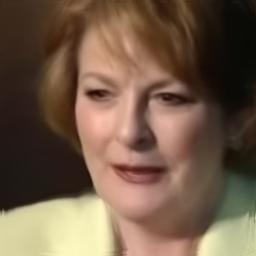}&
        \includegraphics[align=c,width=\wid]{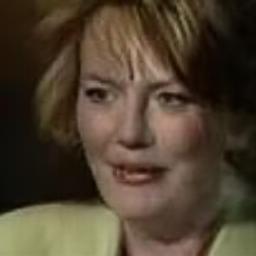}&
        \includegraphics[align=c,width=\wid]{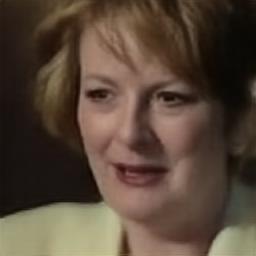}\\
        \centering{\textbf{32}}&
        \includegraphics[align=c,bmargin=0.13cm,width=\wid]{figures_supmat/figure10/1_src.jpg}
        \;&\;
        \includegraphics[align=c,width=\wid]{figures_supmat/figure10/1_gt.jpg}&
        \includegraphics[align=c,width=\wid]{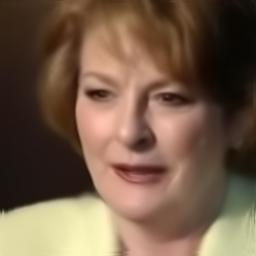}&
        \includegraphics[align=c,width=\wid]{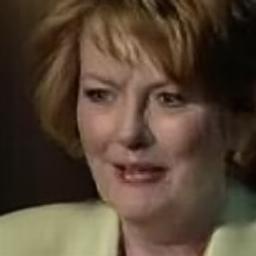}&
        \includegraphics[align=c,width=\wid]{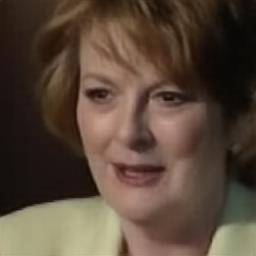}\\
        \centering{\textbf{T}} & \textbf{Source} \;&\; \textbf{Ground truth} & \textbf{X2Face} & \textbf{Pix2pixHD} & \textbf{Ours}
    \end{tabular}
    \caption{First of the extended qualitative comparisons on the VoxCeleb1 dataset. Here, the comparison is carried out with respect to both the qualitative performance of each method and the way the amount of the training data affects the results. The notation for the columns follows \fig{voxceleb1} in the main paper.}\label{fig:voxceleb1suppmatfig1}
\end{figure*}

\begin{figure*}
    \centering    
    \setlength{\wid}{0.179\textwidth}
    \addtolength{\tabcolsep}{-4pt}
    \begin{tabular}{m{0.6cm}c:cccc}
        &
        \includegraphics[align=c,bmargin=0.13cm,width=\wid]{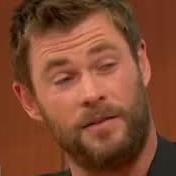}
        \;&\;
        \includegraphics[align=c,width=\wid]{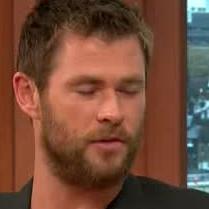}&
        \includegraphics[align=c,width=\wid]{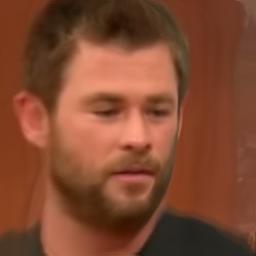}&
        \includegraphics[align=c,width=\wid]{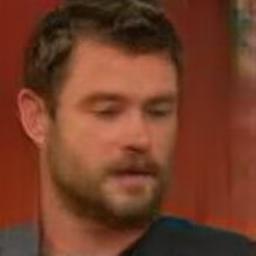}&
        \includegraphics[align=c,width=\wid]{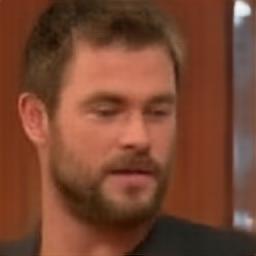}\\
        &
        \includegraphics[align=c,bmargin=0.13cm,width=\wid]{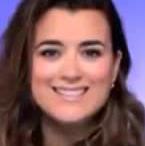}
        \;&\;
        \includegraphics[align=c,width=\wid]{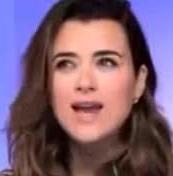}&
        \includegraphics[align=c,width=\wid]{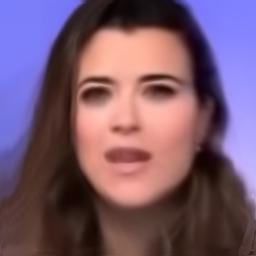}&
        \includegraphics[align=c,width=\wid]{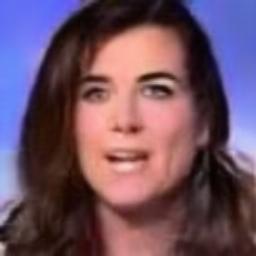}&
        \includegraphics[align=c,width=\wid]{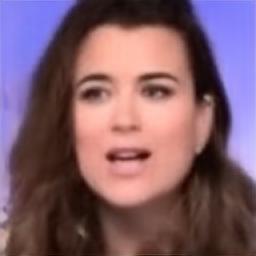}\\
        &
        \includegraphics[align=c,bmargin=0.13cm,width=\wid]{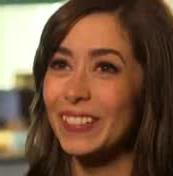}
        \;&\;
        \includegraphics[align=c,width=\wid]{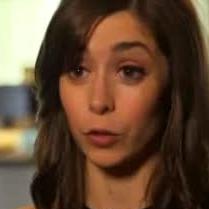}&
        \includegraphics[align=c,width=\wid]{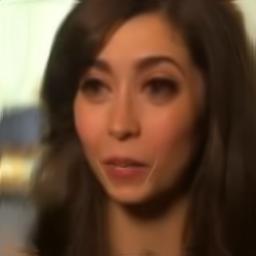}&
        \includegraphics[align=c,width=\wid]{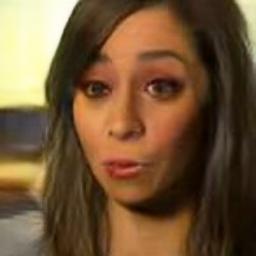}&
        \includegraphics[align=c,width=\wid]{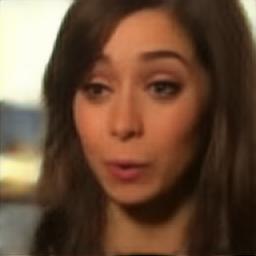}\\
        &
        \includegraphics[align=c,bmargin=0.13cm,width=\wid]{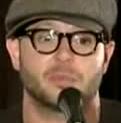}
        \;&\;
        \includegraphics[align=c,width=\wid]{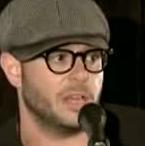}&
        \includegraphics[align=c,width=\wid]{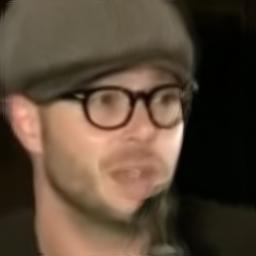}&
        \includegraphics[align=c,width=\wid]{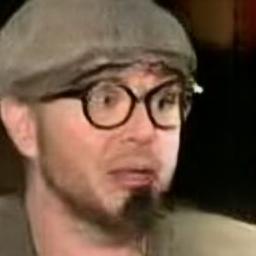}&
        \includegraphics[align=c,width=\wid]{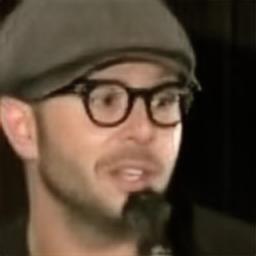}\\
        &
        \includegraphics[align=c,bmargin=0.13cm,width=\wid]{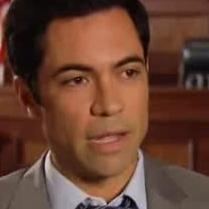}
        \;&\;
        \includegraphics[align=c,width=\wid]{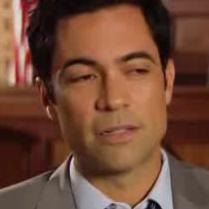}&
        \includegraphics[align=c,width=\wid]{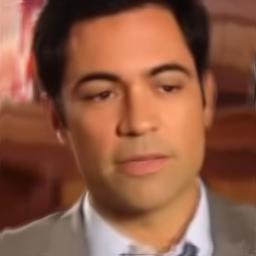}&
        \includegraphics[align=c,width=\wid]{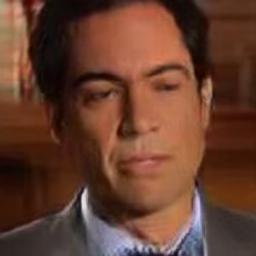}&
        \includegraphics[align=c,width=\wid]{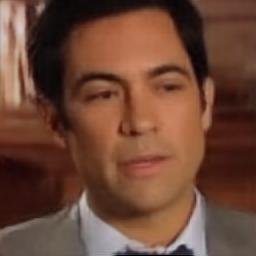}\\
        &
        \includegraphics[align=c,bmargin=0.13cm,width=\wid]{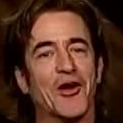}
        \;&\;
        \includegraphics[align=c,width=\wid]{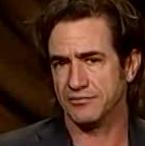}&
        \includegraphics[align=c,width=\wid]{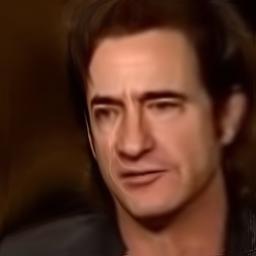}&
        \includegraphics[align=c,width=\wid]{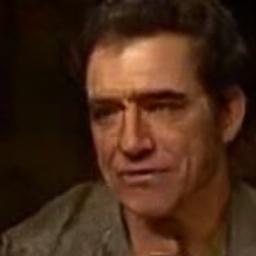}&
        \includegraphics[align=c,width=\wid]{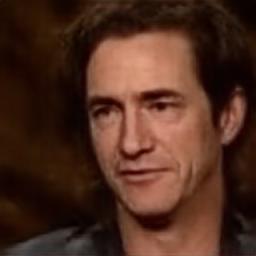}\\
        & \textbf{Source} \;&\; \textbf{Ground truth} & \textbf{X2Face} & \textbf{Pix2pixHD} & \textbf{Ours}
    \end{tabular}
    \caption{Second extended qualitative comparison on the VoxCeleb1 dataset. Here, we compare qualitative performance of the three methods on different people not seen during meta-learning or pretraining. We used eight shot learning problem formulation. The notation for the columns follows \fig{voxceleb1} in the main paper.}\label{fig:voxceleb1suppmatfig2}
\end{figure*}

\begin{figure*}
    \centering    
    \setlength{\wid}{0.179\textwidth}
    \addtolength{\tabcolsep}{-4pt}
    \begin{tabular}{m{0.6cm}c:cccc}
        \centering{\textbf{1}}&
        \includegraphics[align=c,bmargin=0.13cm,width=\wid]{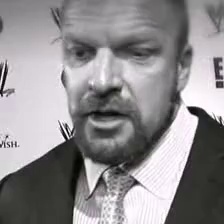}
        \;&\;
        \includegraphics[align=c,width=\wid]{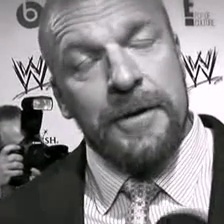}&
        \includegraphics[align=c,width=\wid]{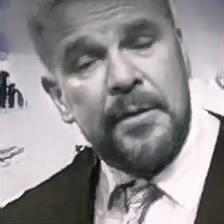}&
        \includegraphics[align=c,width=\wid]{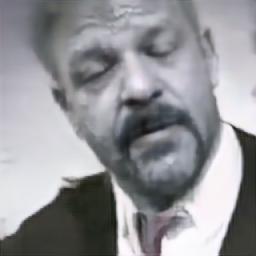}&
        \includegraphics[align=c,width=\wid]{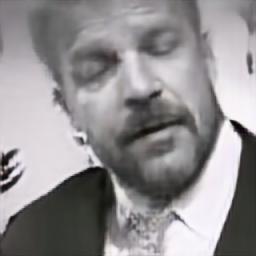}\\
        \centering{\textbf{8}}&
        \includegraphics[align=c,bmargin=0.13cm,width=\wid]{figures_supmat/figure12/0_src.jpg}
        \;&\;
        \includegraphics[align=c,width=\wid]{figures_supmat/figure12/0_gt.jpg}&
        \includegraphics[align=c,width=\wid]{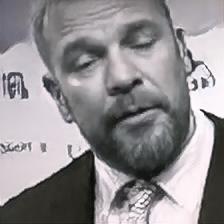}&
        \includegraphics[align=c,width=\wid]{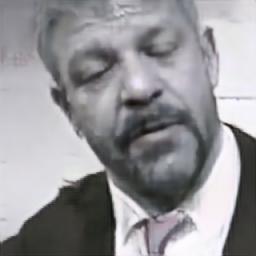}&
        \includegraphics[align=c,width=\wid]{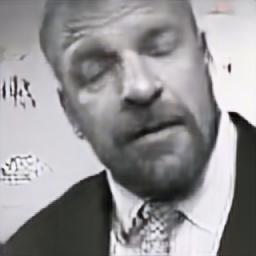}\\
        \centering{\textbf{32}}&
        \includegraphics[align=c,bmargin=0.13cm,width=\wid]{figures_supmat/figure12/0_src.jpg}
        \;&\;
        \includegraphics[align=c,width=\wid]{figures_supmat/figure12/0_gt.jpg}&
        \includegraphics[align=c,width=\wid]{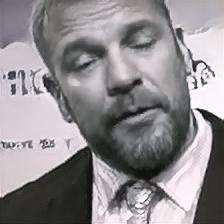}&
        \includegraphics[align=c,width=\wid]{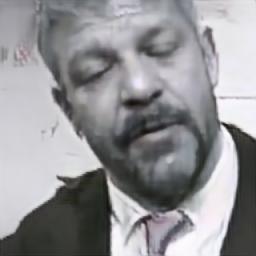}&
        \includegraphics[align=c,width=\wid]{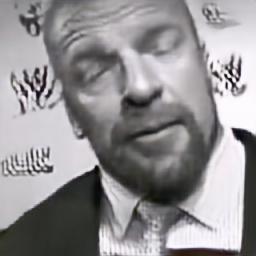}\\
        \centering{\textbf{1}}&
        \includegraphics[align=c,bmargin=0.13cm,width=\wid]{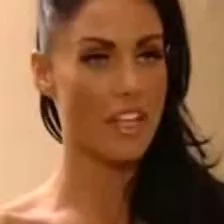}
        \;&\;
        \includegraphics[align=c,width=\wid]{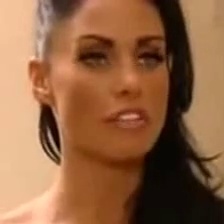}&
        \includegraphics[align=c,width=\wid]{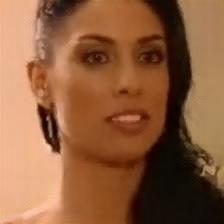}&
        \includegraphics[align=c,width=\wid]{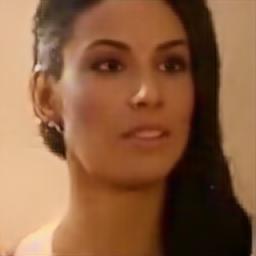}&
        \includegraphics[align=c,width=\wid]{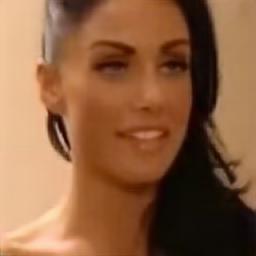}\\
        \centering{\textbf{8}}&
        \includegraphics[align=c,bmargin=0.13cm,width=\wid]{figures_supmat/figure12/1_src.jpg}
        \;&\;
        \includegraphics[align=c,width=\wid]{figures_supmat/figure12/1_gt.jpg}&
        \includegraphics[align=c,width=\wid]{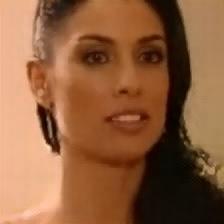}&
        \includegraphics[align=c,width=\wid]{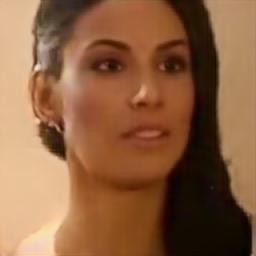}&
        \includegraphics[align=c,width=\wid]{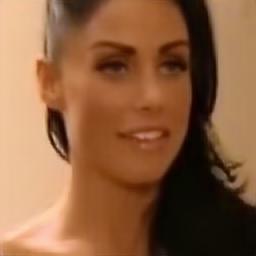}\\
        \centering{\textbf{32}}&
        \includegraphics[align=c,bmargin=0.13cm,width=\wid]{figures_supmat/figure12/1_src.jpg}
        \;&\;
        \includegraphics[align=c,width=\wid]{figures_supmat/figure12/1_gt.jpg}&
        \includegraphics[align=c,width=\wid]{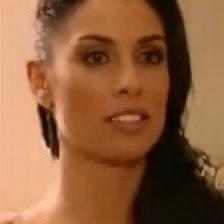}&
        \includegraphics[align=c,width=\wid]{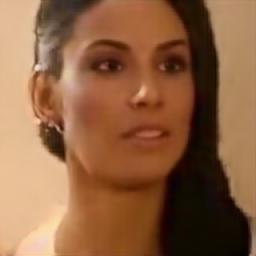}&
        \includegraphics[align=c,width=\wid]{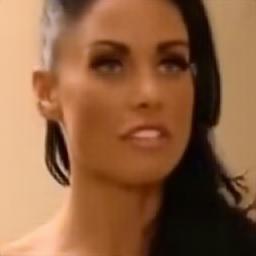}\\
        \centering{\textbf{T}} & \textbf{Source} \;&\; \textbf{Ground truth} & \textbf{Ours-FF} & \textbf{\begin{tabular}{c}Ours-FT\\ before fine-tuning\end{tabular}} & \textbf{\begin{tabular}{c}Ours-FT\\ after fine-tuning\end{tabular}}
    \end{tabular}
    \caption{First of the extended qualitative comparisons on the VoxCeleb2 dataset. Here, the comparison is carried out with respect to both the qualitative performance of each variant of our method and the way the amount of the training data affects the results. The notation for the columns follows \fig{voxceleb2} in the main paper.}\label{fig:voxceleb2suppmatfig1}
\end{figure*}

\begin{figure*}
    \centering    
    \setlength{\wid}{0.179\textwidth}
    \addtolength{\tabcolsep}{-4pt}
    \begin{tabular}{m{0.6cm}c:cccc}
        &
        \includegraphics[align=c,bmargin=0.13cm,width=\wid]{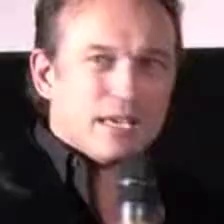}
        \;&\;
        \includegraphics[align=c,width=\wid]{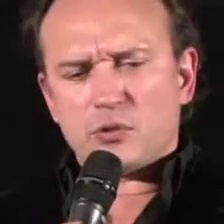}&
        \includegraphics[align=c,width=\wid]{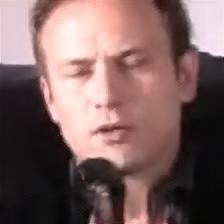}&
        \includegraphics[align=c,width=\wid]{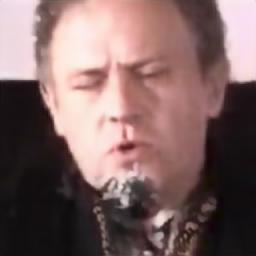}&
        \includegraphics[align=c,width=\wid]{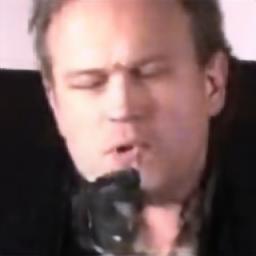}\\
        &
        \includegraphics[align=c,bmargin=0.13cm,width=\wid]{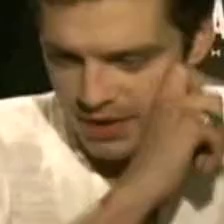}
        \;&\;
        \includegraphics[align=c,width=\wid]{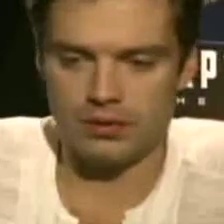}&
        \includegraphics[align=c,width=\wid]{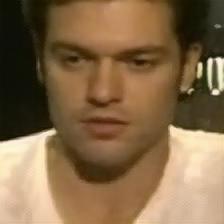}&
        \includegraphics[align=c,width=\wid]{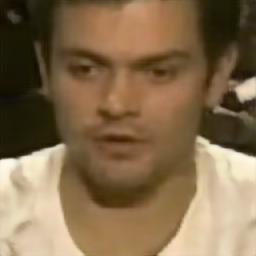}&
        \includegraphics[align=c,width=\wid]{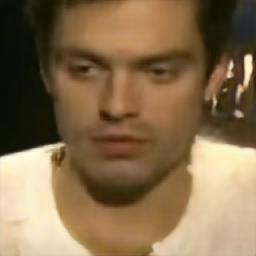}\\
        &
        \includegraphics[align=c,bmargin=0.13cm,width=\wid]{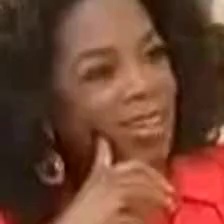}
        \;&\;
        \includegraphics[align=c,width=\wid]{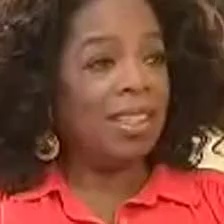}&
        \includegraphics[align=c,width=\wid]{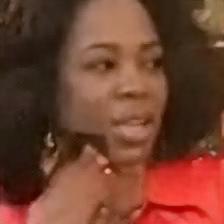}&
        \includegraphics[align=c,width=\wid]{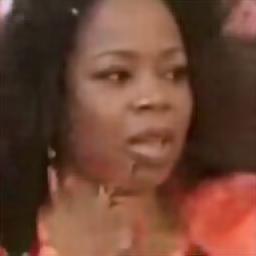}&
        \includegraphics[align=c,width=\wid]{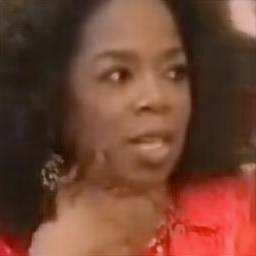}\\
        &
        \includegraphics[align=c,bmargin=0.13cm,width=\wid]{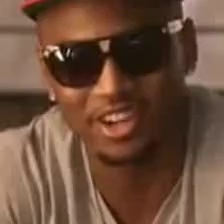}
        \;&\;
        \includegraphics[align=c,width=\wid]{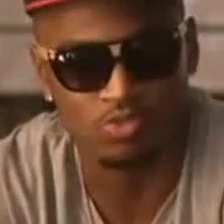}&
        \includegraphics[align=c,width=\wid]{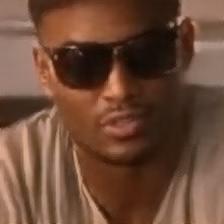}&
        \includegraphics[align=c,width=\wid]{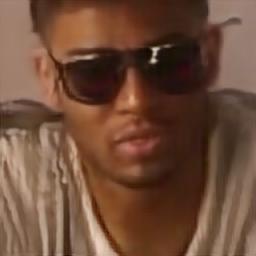}&
        \includegraphics[align=c,width=\wid]{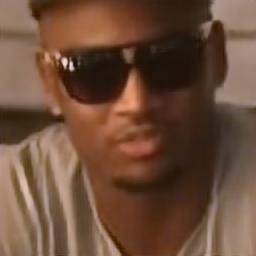}\\
        &
        \includegraphics[align=c,bmargin=0.13cm,width=\wid]{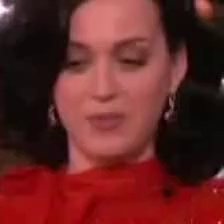}
        \;&\;
        \includegraphics[align=c,width=\wid]{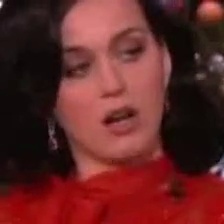}&
        \includegraphics[align=c,width=\wid]{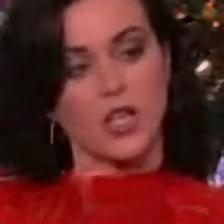}&
        \includegraphics[align=c,width=\wid]{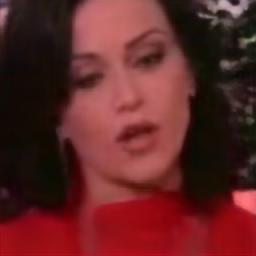}&
        \includegraphics[align=c,width=\wid]{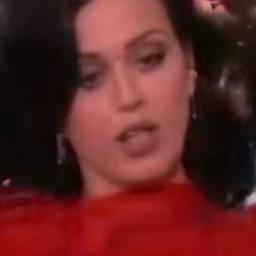}\\
        &
        \includegraphics[align=c,bmargin=0.13cm,width=\wid]{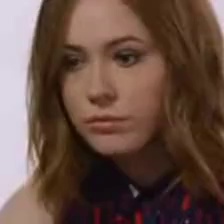}
        \;&\;
        \includegraphics[align=c,width=\wid]{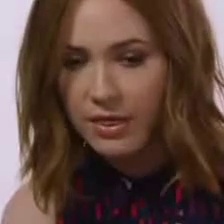}&
        \includegraphics[align=c,width=\wid]{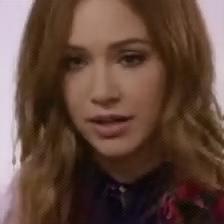}&
        \includegraphics[align=c,width=\wid]{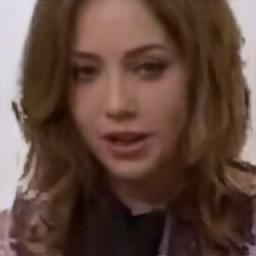}&
        \includegraphics[align=c,width=\wid]{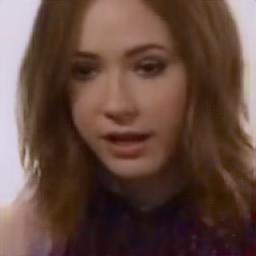}\\
        & \textbf{Source} \;&\; \textbf{Ground truth} & \textbf{Ours-FF} & \textbf{\begin{tabular}{c}Ours-FT\\ before fine-tuning\end{tabular}} & \textbf{\begin{tabular}{c}Ours-FT\\ after fine-tuning\end{tabular}}
    \end{tabular}
    \caption{Second extended qualitative comparison on the VoxCeleb2 dataset. Here, we compare qualitative performance of the three variants of our method on different people not seen during meta-learning or pretraining. We used eight shot learning problem formulation. The notation for the columns follows \fig{voxceleb2} in the main paper.}\label{fig:voxceleb2suppmatfig2}
\end{figure*}